\documentclass[11pt]{article}
\usepackage{coling2020}
\usepackage{times}
\usepackage{url}
\usepackage{latexsym}
\usepackage[dvipdfmx]{graphicx}
\usepackage[subrefformat=parens,labelformat=parens]{subfig}
\usepackage{setspace}
\usepackage{booktabs}
\usepackage{comment}
\usepackage{amsmath,amssymb}
\usepackage{tabularx}
\usepackage{floatrow}
\usepackage{wrapfig}
\usepackage{enumerate}
\usepackage{enumitem}
\usepackage{bm}
\usepackage{placeins}
\usepackage{soul}
\usepackage{color}
\usepackage{multirow}

\newcommand{\cupp}{u}
\newcommand{\lstm}{{\mathrm{LSTM}}}
\newcommand{\PN}{\scalebox{0.95}{\textsf{Paren}}}
\newcommand{\TG}{\scalebox{0.95}{\textsf{Tag}}}
\newcommand{\PNW}{\scalebox{0.95}{\textsf{Paren+W}}}
\newcommand{\TGW}{\scalebox{0.95}{\textsf{Tag+W}}}
\newcommand{\WD}{\scalebox{0.95}{\textsf{Words}}}

\def\vec#1{\overrightarrow{#1}}

\colingfinalcopy 

\title{How LSTM Encodes Syntax: Exploring Context Vectors and\\
Semi-Quantization on Natural Text}

\author{Chihiro Shibata  \\
  Tokyo University of Technology \\
  {\tt shibatachh@stf.teu.ac.jp} 
  \And
  Kei Uchiumi \\
  Denso IT Laboratory  \\
  {\tt kuchiumi@d-itlab.co.jp}
  \AND
  Daichi Mochihashi \\
  The Institute of Statistical Mathematics \\
  {\tt daichi@ism.ac.jp }
}

\date{}

\begin{document}
\maketitle
\vspace{1em}
\begin{abstract}
Long Short-Term Memory recurrent neural network (LSTM) is widely used and
known to capture informative long-term syntactic dependencies.
However, how such information are reflected in its internal vectors for natural
text has not yet been sufficiently investigated.
We analyze them by learning a language model where syntactic structures are implicitly given.
We empirically show that
the context update vectors, {\it i.e.} outputs of internal gates, 
are approximately quantized to binary or ternary values to help
the language model to count the depth of nesting accurately,
as \newcite{Suzgun:19} recently show for synthetic Dyck languages.
For some dimensions in the context vector, we show that
their activations are highly correlated with the depth of phrase structures, such as VP and NP.
Moreover, with an $L_1$ regularization,
we also found that it can accurately predict whether a word is inside a phrase structure or not
from a small number of components of the context vector.
Even for the case of learning from raw text, context vectors are shown to still
correlate well with the phrase structures.
Finally, we show that natural clusters of the functional words and the part of speeches
that trigger phrases are represented in a small but principal subspace of the context-update 
vector of LSTM.
\end{abstract}

\section{Introduction}
\label{intro}

%
%
\blfootnote{
    %
    %
    %
    %
    %
     \hspace{-0.65cm}  
     This work is licensed under a Creative Commons 
     Attribution 4.0 International License.
     License details:
     \url{http://creativecommons.org/licenses/by/4.0/}.
}

\begin{figure}[t]
  {\tabcolsep=0pt
  \begin{tabular}{c@{\kern-1.5em}cc}
  \multicolumn{1}{c}{\includegraphics[height=0.1333\hsize]{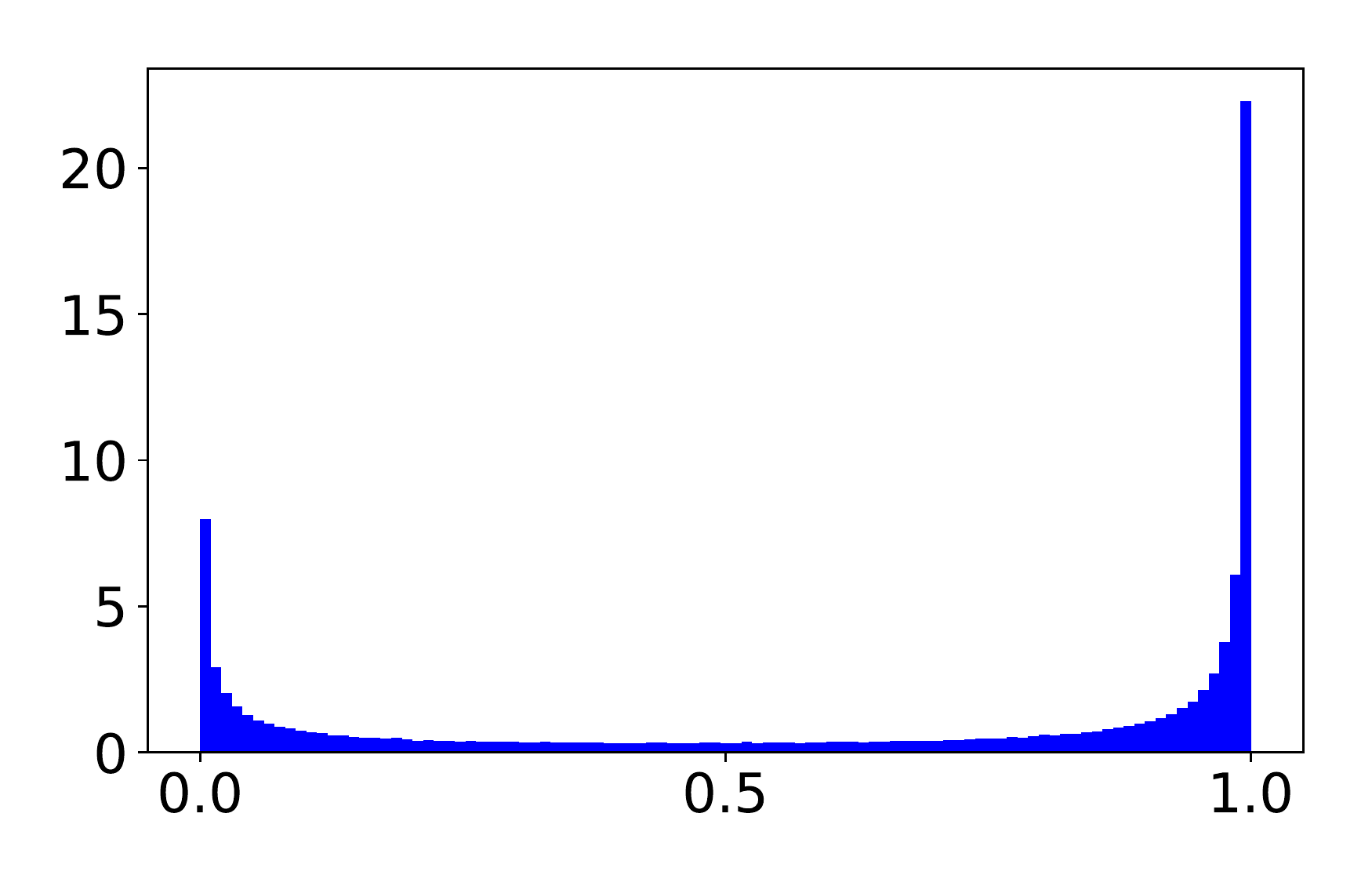}} &
  \multicolumn{2}{c}{\includegraphics[height=0.1333\hsize]{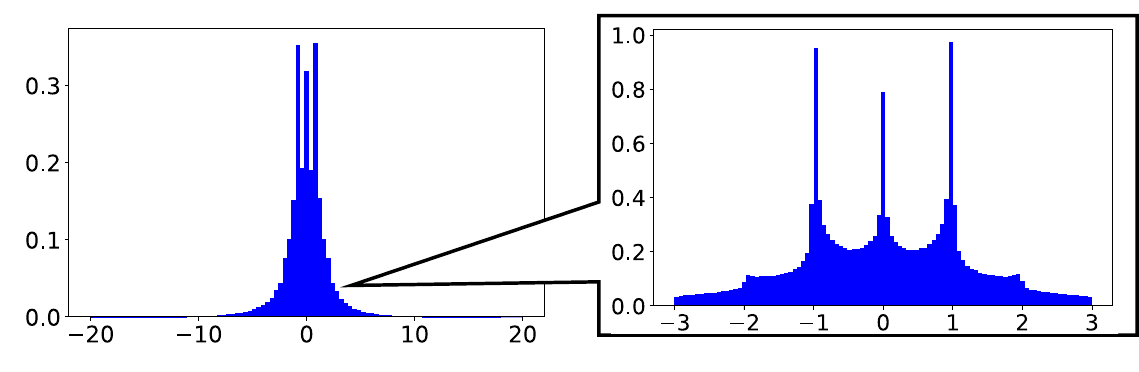}}
  \\[-0.2em]
  \multicolumn{1}{l}{(a) forget vector $\bm{f}$} & 
  \multicolumn{2}{c}{(b) context vector $\bm{c}$}
  \\
  \multicolumn{3}{c}{%
  \includegraphics[width=0.6\hsize]{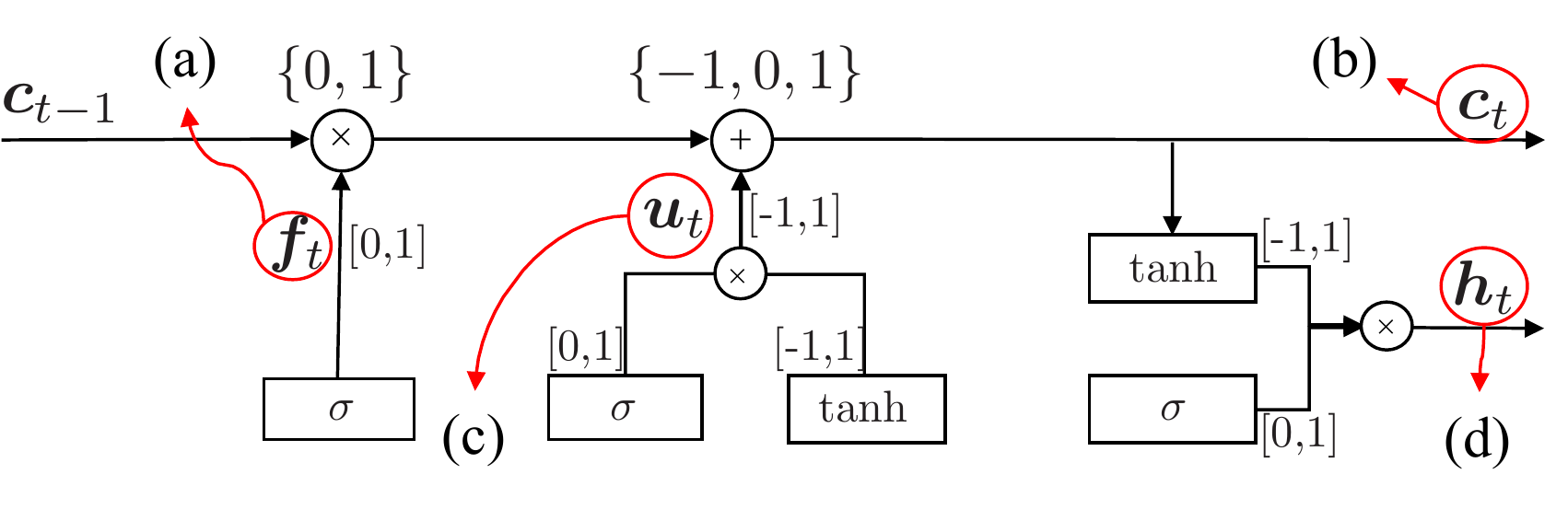}%
  }
  \\[-0.5em]
  \multicolumn{2}{c}{\includegraphics[width=0.2\hsize]{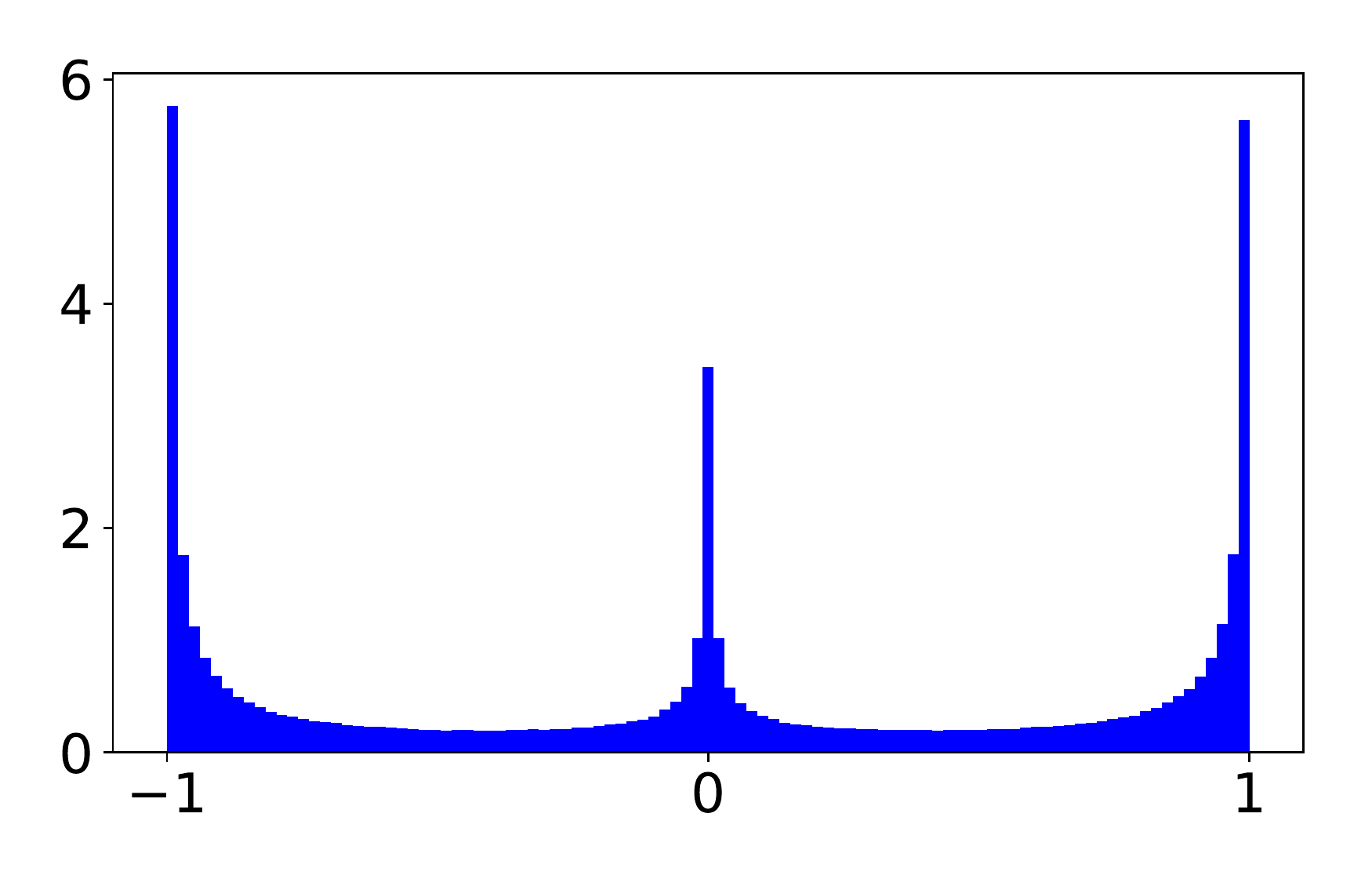}} &
  \includegraphics[width=0.2\hsize]{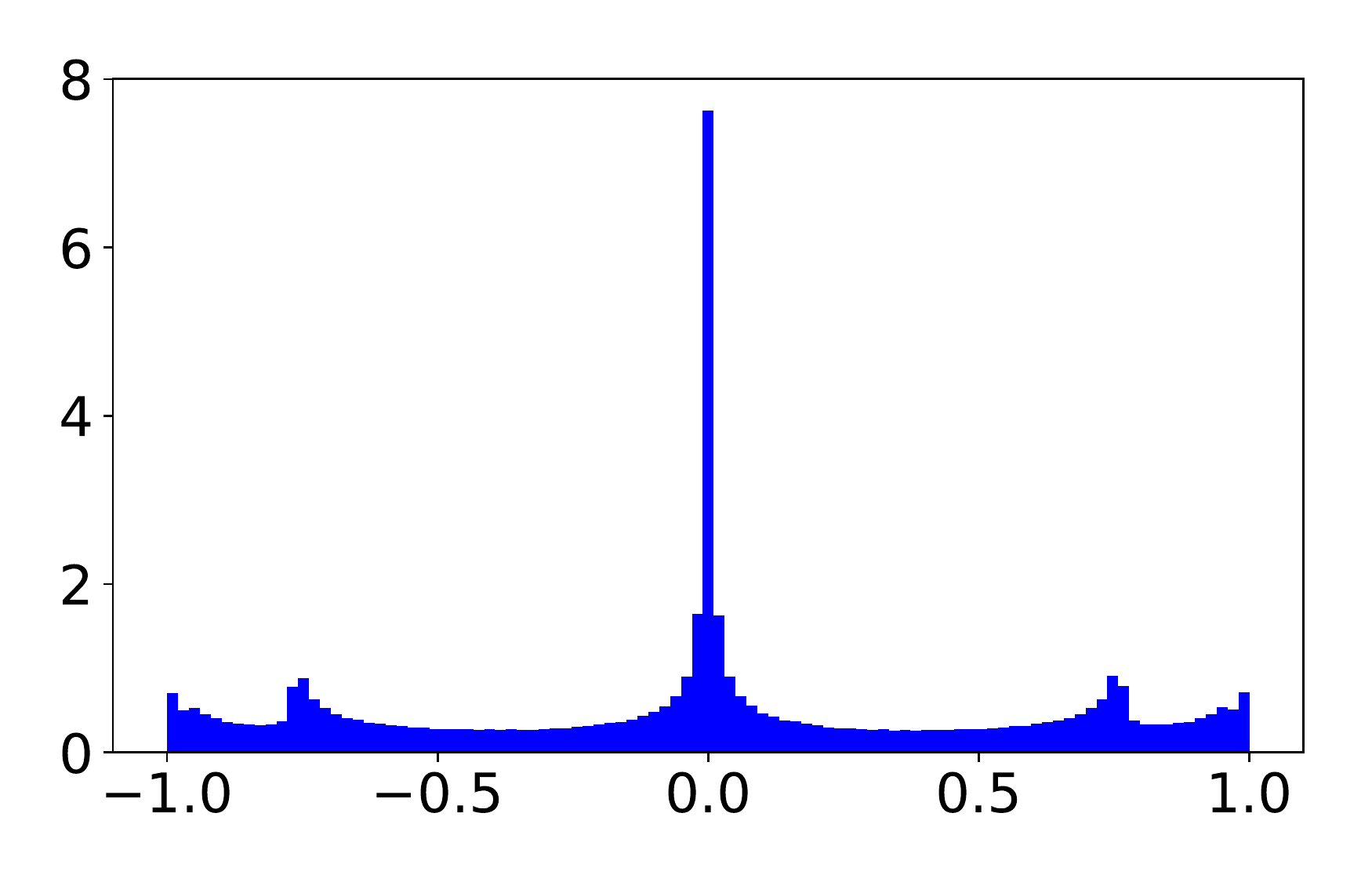}
  \\[-0.2em]
  \multicolumn{2}{c}{(c) context-update vector $\bm{\cupp}$} &
  (d) output-vector $\bm{h}$
  \end{tabular}
  }
    \vspace{-0.5em}
    \caption{Structure of LSTM with
    distributions of elements of the context vector and the surrounding internal vectors. 
    (a) elements of forget vector $\bm{f}$ are nearly binarized to $\{0,1\}$. 
    (b) context-update vector $\bm{\cupp}$ is ternarized to $\{-1,0,1\}$. 
    (c) context vector $\bm{c}$ has several peaks on integers. 
    (d) output vector $\bm{h}$ has peaks at around $\{0,\pm 0.75, \pm 1\}$.
    }
    \label{fig:internal}
  \end{figure}
  LSTM~\cite{Hochreiter:97} is one of the most fundamental architectures that support
  recent developments of natural language processing. 
  It is widely used for building accurate language models by controlling 
  the flow of gradients and tracking informative long-distance dependencies
  in various tasks such as machine translation, summarization and text generation~\cite{wu2016googles,see2017point,fukui2016}.
  While attention-based models such as
  Transformer~\cite{Ashish:17} and BERT~\cite{devlin19bert} and their extensions
  are known to encode syntactic information within it~\cite{Clark:19},
  some studies show that LSTMs are still theoretically superior and empirically competitive in terms of ability to capture syntactic dependency
  \cite{Hahn:19,Dai:19}.
  \newcite{Gongbo:18} and \newcite{Mahal:19} empirically demonstrate that Transformers do not outperform LSTM with respect to tasks to capture syntactic information.

  Recent empirical studies attempt to explain deep neural network models
  and to answer the questions such as how RNNs capture the long-distance dependencies,
  and how abstract or syntactic information is embedded inside deep neural network models \cite{Kuncoro:18,Karpathy:16,Blevins:18}.
  They mainly discuss the extent to which the RNN acquires syntax by comparing
  experimental accuracy on 
  some syntactic structures, such as number agreements (see Section~\ref{sec:related-work}
  for details).
  Some studies also investigate in which vector spaces and layers a specific syntactic information is captured \cite{Liu:18a,Liu:19}.
  Lately, \newcite{Suzgun:19} 
  trained LSTM on Dyck-\{1,2\} formal languages, and showed that it can emulate counter machines.  
  However, no studies have shed light on the inherent mechanisms of LSTM and their relevance to its internal representation in actual text.
  
   \mbox{\newcite{Weiss:18a}} theoretically showed that under a realistic condition, 
  the computational power of RNNs are much more limited than previously envisaged, 
  despite of the fact that RNNs are Turing complete \mbox{\cite{Chen:18}}. 
  On the other hand, they also showed that LSTM is stronger than the simple RNN
  (SRNN) and GRU owing to the counting mechanism LSTM is argued to possess.
  Following these results, \newcite{Merrill:19}
  introduces an inverse temperature $\theta$ into the sigmoid and $\tanh$ functions
  and taking limits as $\theta \to \infty$, and thus assumes that all  gates
  of LSTM are asymptotically quantized: e.g. 
  $\lim_{\theta\to\infty}\sigma(\theta x ) \in\{0,1\}$ and 
  $\lim_{\theta\to\infty}\sigma(\theta x )\tanh(\theta y) \in \{-1,0,1\}$. 
  Under the above assumption,  it shows LSTMs work like counter machines, or more precisely, the expressiveness of LSTMs is asymptotically equivalent to that of some subclass of counter machines.
  While those results are significant 
  and giving us theoretical clues to understand how LSTMs acquire syntactic representations as their hidden vectors, it is not yet known whether or not similar phenomena occur in models learned from real-world data.
  Regarding this point,
  we show that those quantization actually often happens in real situations and bridge a gap between theories and practical models 
   through statistical analysis of internal vectors of LSTM that are trained from both raw texts and texts augmented by implicit syntactical symbols.
  
  We first explore the behaviors of LSTM language models (LSTM-LMs) and the representation of the syntactic structures by giving linearized syntax trees implicitly. 
  Then, we show that LSTM also acquires a representation of syntactic information
  in their internal vectors even from a raw text, by statistically analyzing the internal vectors corresponding to syntactic functions.
  We empirically show that the representations of
  part of speeches such as NP and VP and syntactic functions that specific words have, both of which often act as syntactic triggers, are acquired in the space of 
  context-update vectors, as well as syntactic dependencies are accumulated in the space of context vectors. 
  \section{Statistics of Internal Vectors of LSTM}

  \subsection{LSTM Language Model}
  
  \newcommand{\aff}{{\mathrm{aff}}}
  \newcommand{\vtm}[2]{ {#1}_{ {#2} } }
  \newcommand{ \vh}[1]{ \vtm{ \bm{h} }{ {#1} } }
  \newcommand{ \vc}[1]{ \vtm{ \bm{c} }{ {#1} } }
  \newcommand{ \vf}[1]{ \vtm{ \bm{f} }{ {#1} } }
  \newcommand{ \vx}[1]{ \vtm{ \bm{x} }{ {#1} } }
  \newcommand{\vup}[1]{ \vtm{\bm{\cupp}}{ {#1} } }
  
  In this study, we consider language models based on one-layer LSTM
  because our aim is to clarify how LSTM captures syntactic structures.
  For a sentence $w_1 w_2\cdots w_n$,  
  as shown in Figure~\ref{fig:internal},
  let $\bm{h}_t$ denote the \emph{output vector} of an LSTM 
  after feeding the $t$-th word $w_t$, $\bm{c}_t$ denote the \emph{context vector}, 
  and $\vec{w_t}$ denote the embedding of the word $w_t$.
  Let $\lstm (\bm{c}, \bm{h}, \vec{w}, \Theta)$ be a function of $\bm{c}$, $\bm{h}$ and $\vec{w}$ to determine the next output and context vectors:
  \begin{equation}
  (\vc{t}, \vh{t} )  = \lstm\, (\vc{t-1}, \vh{t-1}, \vec{w_t}, \Theta ),
  \end{equation}
  where $\Theta$ represents the set of parameters to be optimized.
  The language model maximizes the probability of the next word $w_{t+1}$
  given the word sequence up to $t$:
  \begin{equation}
  p(w_{t+1} | w_{1:t}) = p(w_{t+1} | w_t, \vc{t}, \vh{t}) = s( W \vh{t+1} + b ).
  \label{eqn:prediction}
  \end{equation}
  $s( )$ is the softmax function, and
  $W$ and $b$ are a weight matrix and a bias vector, respectively.
  As shown in the equation~\eqref{eqn:prediction}, 
  the history of words up to $t\!-\!1$ does not appear 
  explicitly in the conditional part of the probability.
  The contextual information is represented in some form in the context vector $\vc{t}$ and
  the output vector $\vh{t}$.
  The following standard version is used as the target LSTM architecture 
  among multiple variations~\cite{Greff:17}:
  \vspace{-1em}
  \begin{align}
  \vf{t} &= \sigma( A \vx{t} ) \label{eq:f}\\
  \vup{t} &= \sigma( B \vx{t}  )\odot \tanh( C \vx{t} ) \label{eq:cup} \\
  \vc{t} &= \vf{t} \odot \vc{t-1}  + \vup{t} \label{eq:c}\\ 
  \vh{t} &= \tanh(\vc{t}) \odot \sigma( D \vx{t} ) \label{eq:h}  
  \end{align}
  Here, $\odot$ is an Hadamard (element-wise) product, and
  $\vx{t} $ is the concatenated vector of $\vec{w_t}$, $\vc{t-1}$, $\vh{t-1}$, and $1$.
  $A,B,C,D$ are weight matrices representing affine transformation.
  In this paper, $\bm{\cupp}$ and $\bm{f}$, which are derived from $\vx{}$ by
  equations \eqref{eq:f} and \eqref{eq:cup} to
  directly affect $\bm{c}$, are also analyzed in addition to $\bm{c}$ and $\bm{h}$.
  $\bm{\cupp}$ and $\bm{f}$ are called \emph{context-update} vector and \emph{forget} vector hereinafter. 
  
  \subsection{Internal Vectors are Naturally Quantized}
  
\begin{wraptable}{r}{60mm}
  \vspace{-2\intextsep}
  \begin{tabular}{llll}
      $\bm{h}$  &        $\bm{c}$ &  $\bm{\cupp}$ &   $\theta(\bm{\cupp})$ \\
  \hline
         my &      his &     his &     his\rule{0pt}{0.9em} \\
        his &   mother &      my &      my  \\
     mother &  playing &     the &     the  \\
    husband &     mind &     its &     its  \\
       mind &  husband &     our &    your  \\
       wife &  matters &    your &   their  \\
      their &    party &   their &      's  \\
  \hline
  \end{tabular}
  \caption{Most similar words with ``her'', based on different internal vectors in LSTM.
  $\theta()$ is a discretization by thresholds $\pm 0.9$.
  }
  \label{tab:her}
\end{wraptable}


 The fundamental focus of this study is a natural {\it semi-quantization} of $\bm{f}$, $\bm{c}$, $\bm{\cupp}$, and $\bm{h}$, as the result of learning. 
 First, each element of $\bm{\cupp}$ is approximately quantized, or ternarized, 
 to $\{-1, 0, 1\}$ as shown in Figure~\ref{fig:internal}(c).
  This discretization is a consequence of equation~\eqref{eq:cup}:
  the distribution of the first term is almost concentrated on $0$ and $1$,  and that of the second term is concentrated on $\pm 1$.
  We experimentally confirmed that even if each element of $\bm{\cupp}$ is strictly ternarized by thresholds,
  it does not lose important information. 
  For example, Table~\ref{tab:her} lists the most similar words with the word ``her'' measured by the internal vectors 
  (see Section~\ref{sec:similarities} for details). 
  $\theta(\bm{\cupp})$,
  which is obtained by thresholding $\bm{\cupp}$ by $\pm 0.9$, 
  collects syntactically similar words as appropriately as $\bm{\cupp}$ does.
  
  Each element of $\bm{f}$ is also approximately binarized to $\{0, 1\}$
  as seen in Figure~\ref{fig:internal}(a), where the ratio of $1$ is larger.
  Context-update vector $\bm{\cupp}$ is added to $\bm{c}$ and accumulated as long as the value of $f$ is close to $1$.
  Owning to the effects of such quantization and accumulation, 
  Figure~\ref{fig:internal}(b) shows that
  the distribution of each element of $\bm{c}$ will have peaks on integers.
  
  As we discuss in Section~\ref{sec:RepNest}, 
  this quantization enables the accurate counting of the number of words with syntactic features such as the nesting of parenthesis.
  Note that Figure~\ref{fig:internal} shows the results of learning from the {\it raw text} of Penn Treebank WSJ corpus~\cite{Taylor:03},
  and the characteristics described above do not change even if 
  the parameters such as datasets and the dimensionality of the vectors have been changed.
  
  \section{Hypotheses and Outline of Analyses}
  
  To understand the behavior of LSTM further, we try to answer two kinds of questions:
  (a) {\it what} information is relevant with the syntax, and (b) {\it how} these
  information is correlated with the syntactic behavior.
  In particular, we will examine: (1) which of the internal vectors (i.e. $\bm{h}$, $\bm{c}$, and
  $\bm{\cupp}$) of LSTM highly correlates with the prediction of the phrase structure
  and its nesting (Sections 5.1 and 5.2),
  and (2) how well these internal vectors or some subsets of 
  their dimensions can predict the syntactic structures (Section 5.3).
  Since recognition of syntax inevitably requires recognition of the part-of-speech
  for each word, we also investigate: (3) how the contextual part-of-speech is
  represented in the internal vectors of the LSTM, and how the differences between them
  can be captured using PCA (Section 6).
  
  
  \section{Target Datasets and Learned Models}
  \subsection{Configuration of Datasets}
  We use sentences with syntax trees in Peen Treebank Wall Street Journal (PTB-WSJ) corpus~\cite{Marcus:94,Taylor:03} as data for training and testing.
  We chose evenly spaced $10\%$ of data for testing.
  Phrase structures are linearized and inserted into (or replaced with) sentences 
  as auxiliary tokens in several manners as follows:
  \begin{description}[itemsep=-5pt,topsep=4pt]
      \item[\mbox{\PN}] consists of only `(' and `)' without words,
      \item[\PNW] consists of `(' and `)' and words,
      \item[\TG] consists of `({\it T}' and `{\it T})' without words, where 
      {\it T} represents a tag in Penn Treebank,
      \item[\TGW] consists of `({\it T}' and `{\it T})' and words,
      \item[\WD] is just a set of raw words.
  \end{description}
  For example, a sentence in the original data
  ``(NP (DT a) (JJ nonexecutive) (NN director))''
  is converted to
  ``(() () ())'' in {\PN} and ``(NP (DT DT) (JJ JJ) (NN NN) NP)'' in \TG.
  The latter needs some attention; here, each space-separated
  token such as ``('', ``(NP'', or ``JJ)'' 
  is considered as a single word.
  The size of the vocabulary in {\PN} and {\TG} is 2 and 140, respectively.
  For {\PNW} and {\TGW}, less frequent words
  were replaced by their parts of speech so that the total number of words was less than 10,000.
  Additionally, we also included a small experiment using Lisp programs:
  in particular, we used {\tt slib} standard library of {\tt scheme} and conducted experiments 
  under the scenarios {\PN} and {\TG} to show that LSTM also works similarly for other
  ``languages'' other than WSJ. 
  Note that in the all scenarios above, LSTM does not know the correspondence between
  ``({\it T}'' and ``{\it T})'' for each tag {\it T} in advance, because these
  auxiliary ``words'' are simply converted to integers like any other words and fed to
  LSTM.
  Therefore, syntactic supervision in our experiments is not complete but only hinted.
  
  \floatsetup[figure]{subcapbesideposition=center}
  {\textfloatsep=0pt
  \begin{figure*}[t!]
      \centering
      \subfloat[][\PNW]{
          \includegraphics[width=0.32\columnwidth]{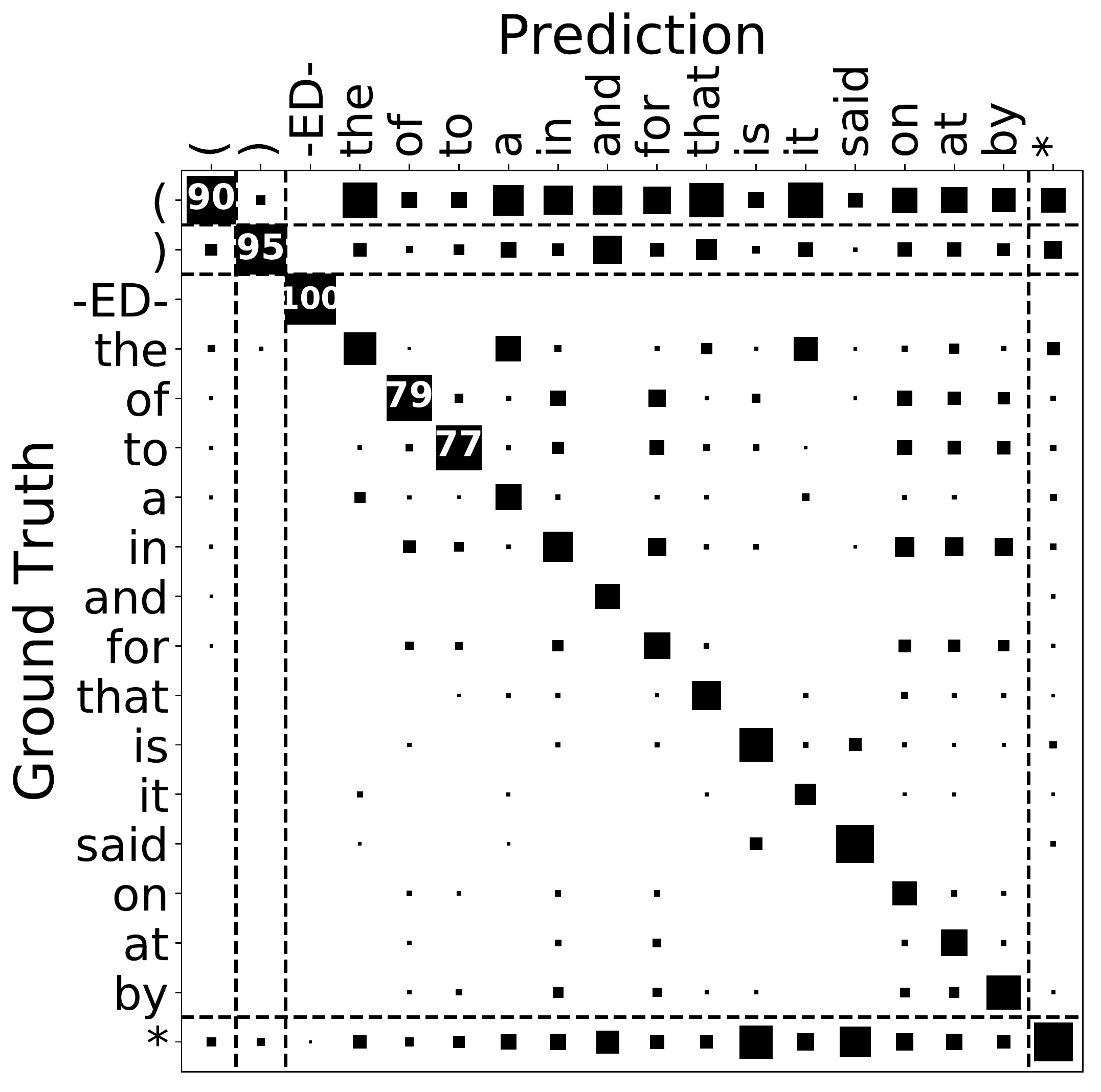}
          \label{fig:c3}
      }
      \subfloat[][\TG]{
          \includegraphics[width=0.32\columnwidth]{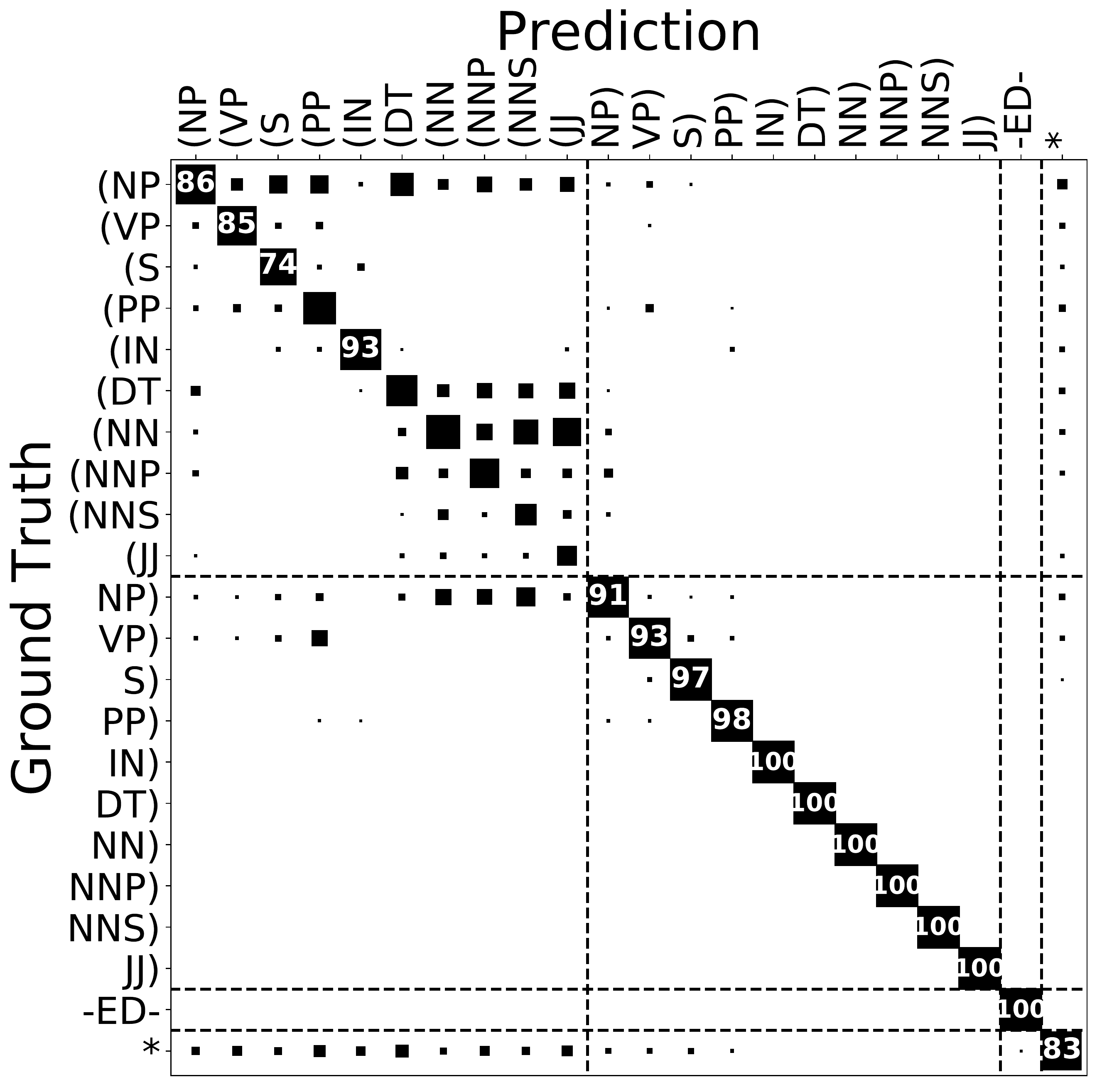}
          \label{fig:c2}
      }
      \subfloat[][\TGW]{
          \includegraphics[width=0.32\columnwidth]{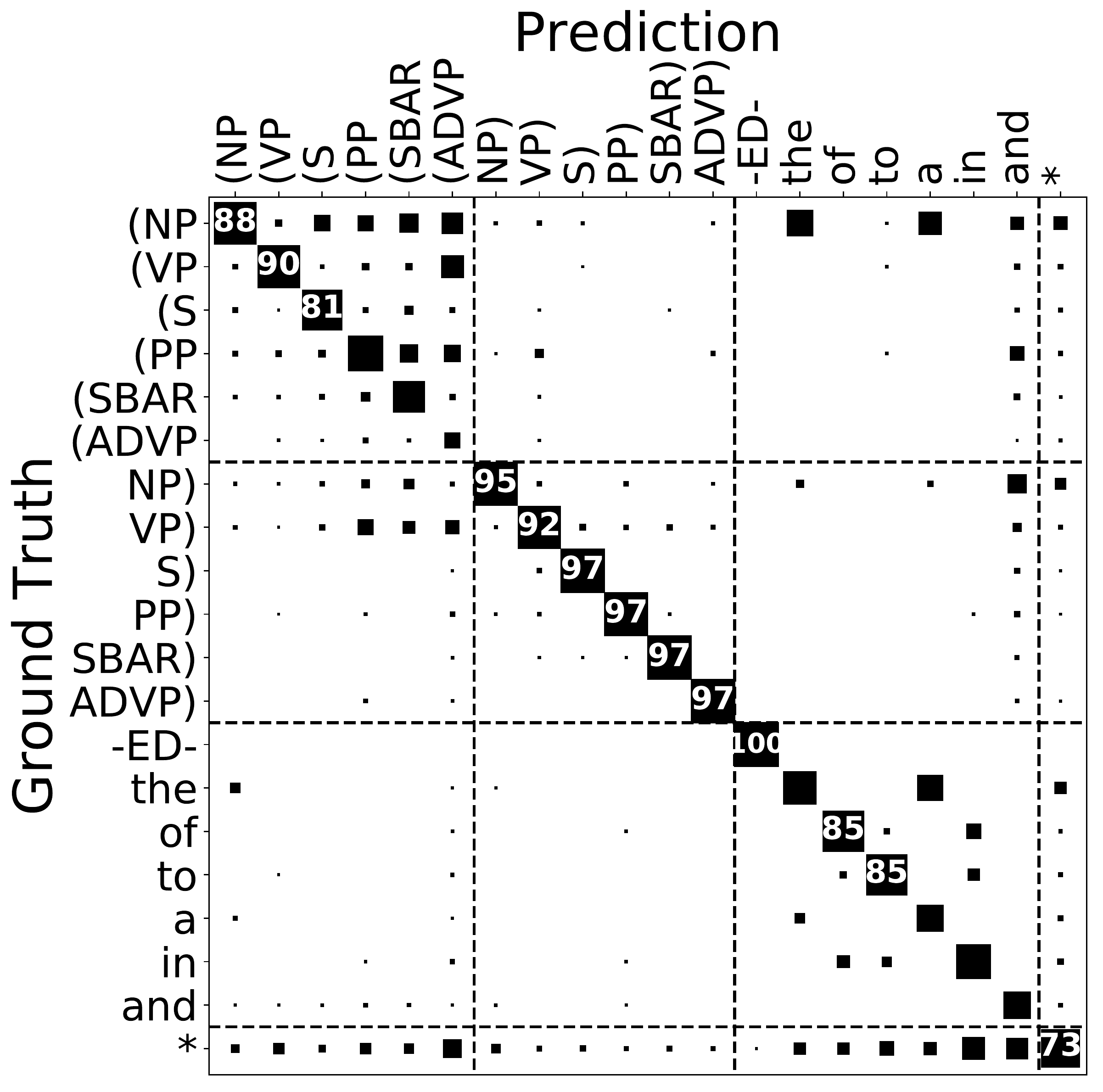}
          \label{fig:c4}
      }
      \vspace{1em}
      \caption{Confusion matrices of next word prediction on test data.
      Figure (a)--(c) correspond to the datasets {\PNW}, {\TG}, and {\TGW};
      dashed lines show the groups of tokens.
      The number within a cell shows the precision as a percentage. 
      Only frequent words are shown and infrequent words are collectively denoted by `*'. \\[-1.2em]
      }
      \label{fig:call}
      \vspace{-1.5em}
  \end{figure*}
  }
  
  \subsection{Learning Models}

  The simplest architecture for LSTM language model is employed, which is
  composed of a single LSTM layer with a word embedding and a softmax layer.
  The size of the word embedding vectors and the internal vectors are determined according to 
  the size of the vocabulary: 100 for \PN, 200 for \TG, 
  and 1,000 for \PNW, \TGW, and {\WD} because they include actual words.
  We used Adam~\cite{Kingma:15} for optimization, where hyperparameters such as the step size are the same as \cite{Kingma:15}.
  After 20 epochs of training, a model that has the best accuracy for test data among all epochs
  is chosen for analysis.
  For the case of {\WD}, 
  Dropout is applied to emulate the actual usage of LSTM.
  The rates of Dropout are set to $0.2$ and $0.5$ for the embedding and output vectors, respectively. 
  
 \subsection{Prediction Accuracies}
 
 \begin{wraptable}[11]{r}{70mm}
      \centering
      \vspace{-0.2em}
      \begin{spacing}{0.95}
      \begin{tabular}{l|c|c|c|c}
          Dataset & BOP          & EOP     &     EOS     & Words   \\
          \hline
          \PN\rule{0pt}{0.9em}  &  0.77    &      0.87    & {\bf 1.00}    & --    \\
          \PNW    &      0.90   &  0.96    &     {\bf 1.00}    & 0.78   \\
          \TG     &      0.87   &  0.93    &     {\bf 1.00}    & --    \\
          \TGW    &      0.89   &  0.96    &     {\bf 1.00}    & 0.86   \\
          \WD     &      --     &  --      &     --      & 0.49  \\
           \hline
      \end{tabular}
      \end{spacing}
      \caption{Micro-averaged precision of prediction for the beginnings of phrases (BOPs), ends of phrases (EOPs),  end of sentence (EOS), and raw words.
      }
      \label{tab:aux_accs}
  \end{wraptable} 
  We compare the accuracy of predicting the next word among different datasets to phenomenologically confirm 
  the acquisition of phrase structures.
  As shown in Table~\ref{tab:aux_accs},
  the end of sentence (EOS) is predicted by LSTM almost perfectly in terms of both precision and recall
  for all datasets except for {\WD}.
  Because EOS occurs in a sentence if and only if the numbers of \mbox{`({\it T}'} and \mbox{`{\it T})'} 
  are equal for all $T$,
  we can conjecture that the LSTM model counts the balance and the nesting of them accurately.
  
  In Figure~\ref{fig:call}, the groups of Beginning of Phrase
  (BOP, i.e. ``({\it T}'' for a tag {\it T}) and End of Phrase (EOP, i.e. ``{\it T})'') 
  are separated by the dashed lines. 
  We can see that BOP and EOP are correctly classified across groups (Figure~\subref*{fig:c2},~\subref*{fig:c4}).
  Furthermore, each EOP is rarely misclassified to another EOP. 
  This implies that not only the balance of the numbers of `{\it (T}'and `{\it T)}' is completely learned,
  but their order of appearance is also learned quite accurately.
  Comparing Figure~\subref*{fig:c4} to \subref*{fig:c2}, we can see that 
  the precisions for BOP and EOP are improved by including intervening words.
  Similarly, the precisions for the words are also improved
  by including the auxiliary tokens (Table~\ref{tab:aux_accs}).
  These are because the existence of words will serve as a clue to predict phrase structures,
  and vice versa.
  
  \section{Representation of Syntactic Structures}
  
  After these investigations on LSTM, next we will examine
  how each tag of the phrase structure and 
  the depth of the nesting are embedded in its internal vectors.
  
  \subsection{Depth of Nested Phrases}
  
    \begin{figure}[!b]
      \begin{tabular}{cc}
        \begin{minipage}[b][50mm][s]{0.42\columnwidth}
        \vfill
        \subfloat[][\PN]{
          \includegraphics[width=0.5\columnwidth,trim=20 20 20 10,clip]
          {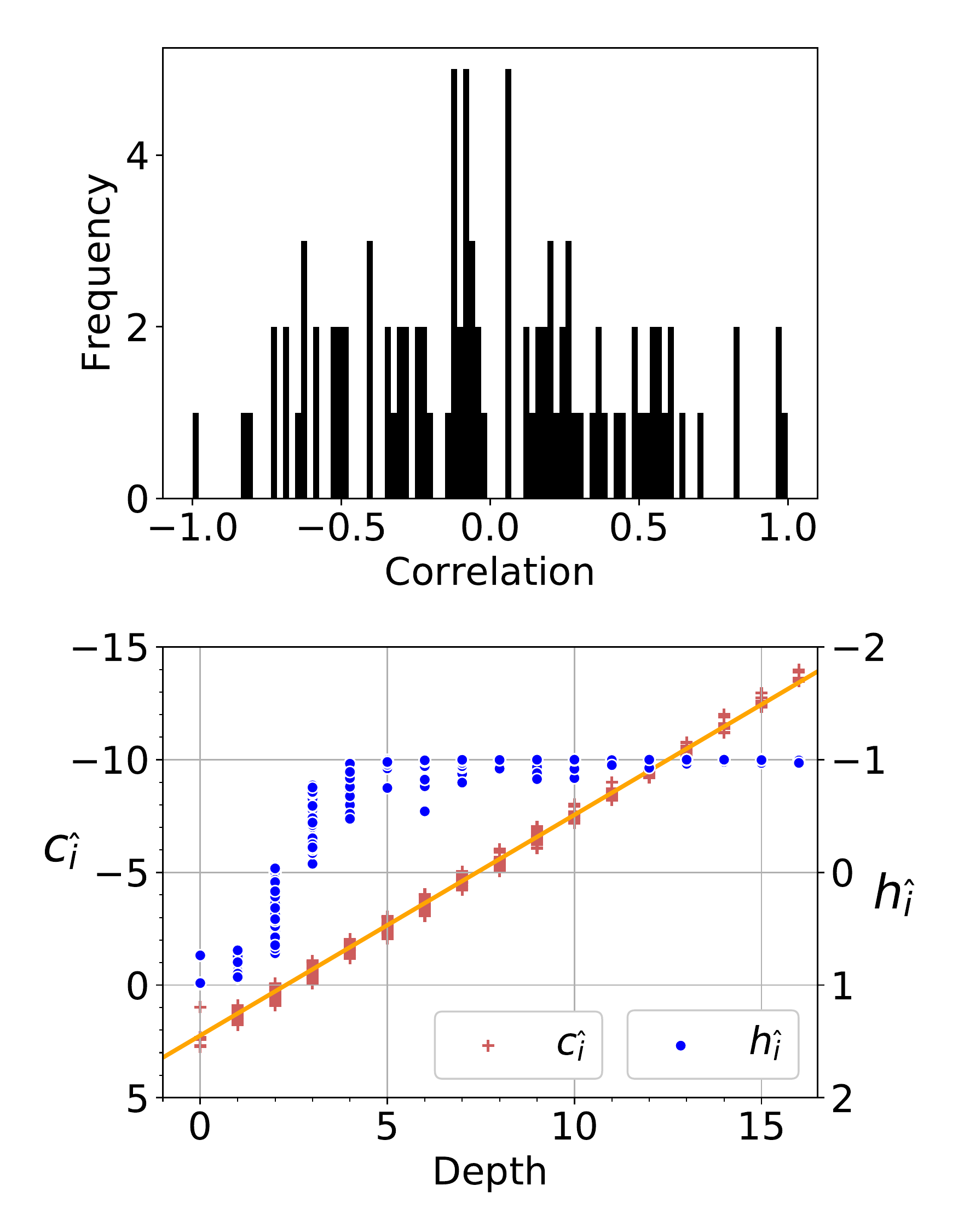}
          \label{fig:cor1}
        }
        \subfloat[][\PNW]{
          \includegraphics[width=0.5\columnwidth,trim=20 20 20 10,clip]
          {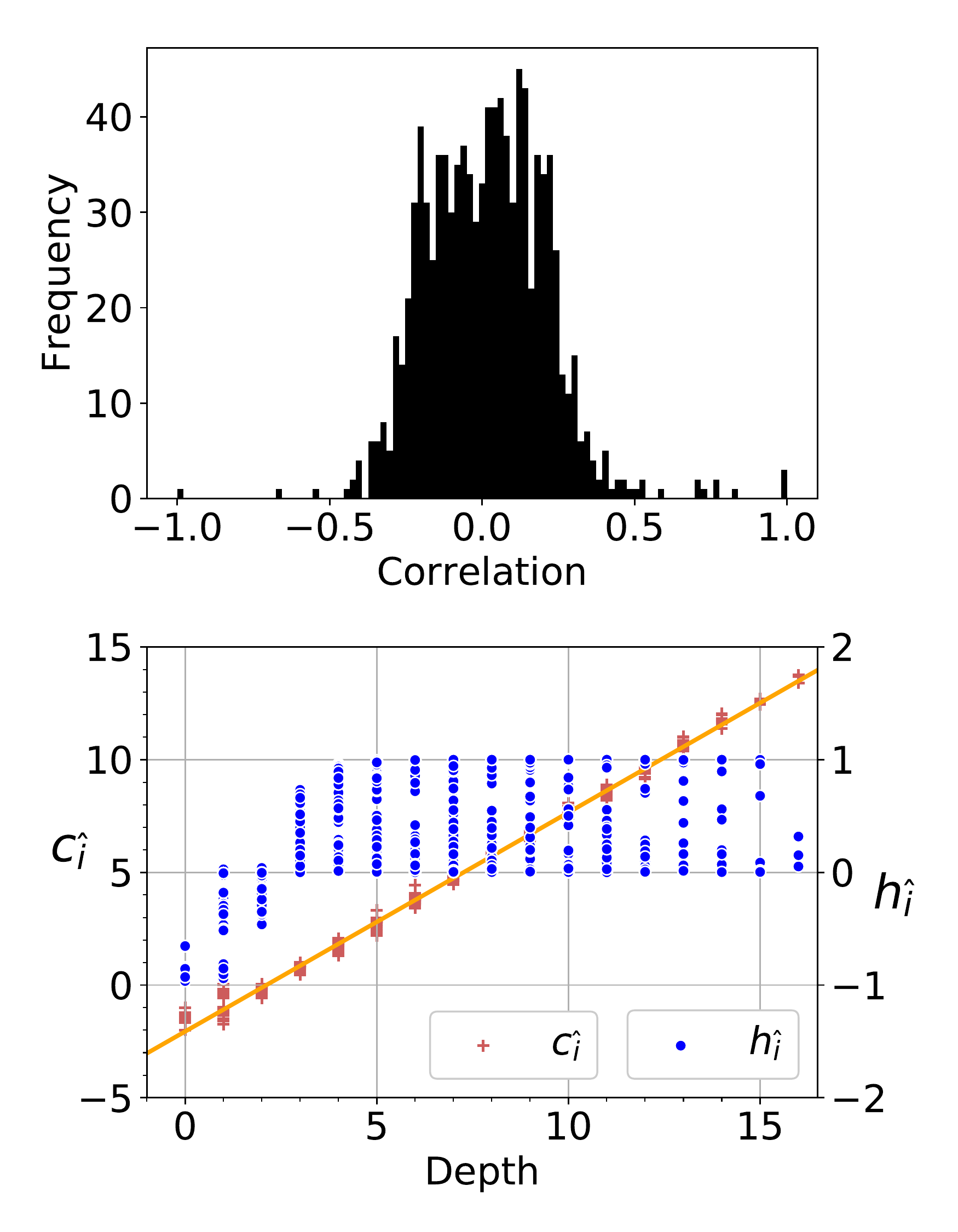}
          \label{fig:cor3}
        }
      \end{minipage}
       &
        \begin{minipage}[b][50mm][s]{0.52\columnwidth}
        \vfill
      \subfloat[][\PN]{
          \includegraphics[width=\columnwidth, trim=60 50 40 40,clip]
          {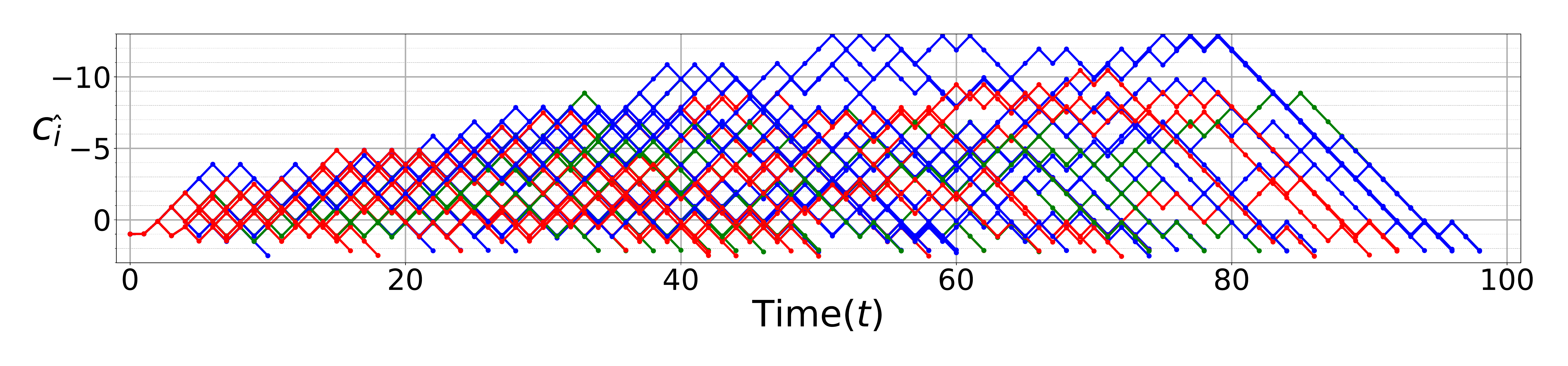}
          \label{fig:grid}
      }\\
      \subfloat[][Lisp]{
          \includegraphics[width=\columnwidth, trim=60 50 40 40,clip]{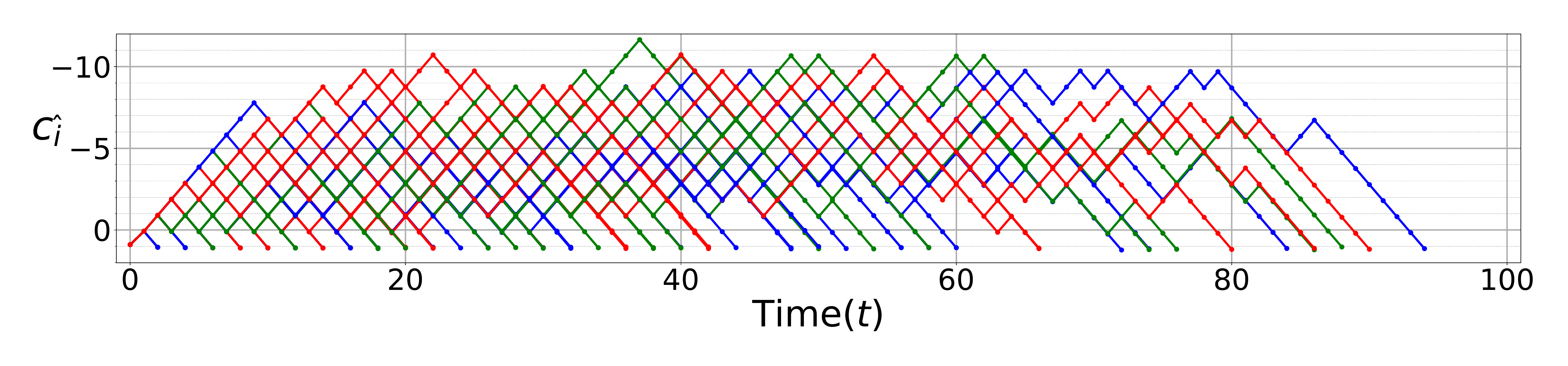}
          \label{fig:grid-slib}
      }
      \end{minipage}
      \end{tabular}
    \caption{(a) and (b):
      (Upper) histogram of correlations with the nesting for each element of $\bm{c}$. 
      (Lower) plots of the activations of 
      $\bm{c}$ (red) and $\bm{h}$ (blue) for the dimension of highest correlation. 
      (c) Value of $c_{\hat{\imath}}$ as a function of time in each sentence.
      Each trajectory represents a sentence in the test data.
      Trajectories are colored so that each one is easily distinguished.
      (d) same as (c) on Lisp programs.
      We can see that
      a mesh structure with the step height of approximately $1$ emerges
      in spite of the continuous space of $\bm{c}$.
      }
    \label{fig:grid1}
  \end{figure}
  
  We first examine the correlation coefficients between the depth of nesting and the value of each dimension
  of the context vector $\bm{c}$.
  Results are shown
  in the upper half of Figure~\subref*{fig:cor1} and \subref*{fig:cor3} for
  {\PN} and {\PNW}, respectively.
  There are some dimensions whose correlation are very high; 
  $0.9969$, $0.9978$, and $0.9995$ for {\PN}, {\PNW}, and Lisp, respectively. 
  Let $\hat{\imath}$ denote the dimension such that this correlation is maximized. 
  As Figure~\subref*{fig:cor1} and \subref*{fig:cor3} show, we can see that the depth of the nesting 
  linearly correlates with $c_{\hat{\imath}}$ and
  almost equals to $|c_{\hat{\imath}}|-\alpha$, with some constatnt $\alpha$.
  In contrast, the values of $h_{\hat{\imath}}$ in $\bm{h}$ are scattered;
  especially for {\PNW}, $|h_{\hat{\imath}}|$ does not converge to $1$ and has a large variance between $0$ and $1$. 
  The first term 
  in the right-hand side of equation~\eqref{eq:h} leads to this variance
  because the second term is nearly $1$ or $-1$ when the nesting is deep.
  
  In Figure~\subref*{fig:grid}, we randomly choose dozens of sentences from the test data
  whose lengths are less than 100,
  and plot the values of $c_{\hat{\imath}}$ as time proceeds.
  We can see that a mesh structure is obtained with the step height of nearly $1$
  in spite of the continuous space of $\bm{c}$.
  This is because, as described in Section 2.2, 
  the context-update vectors $\bm{\cupp}$ are approximately quantized so that 
  $\cupp_{\hat{\imath}}$ is almost binarized to $\pm 1$.
  In addition, the end points of the graphs have values of approximately $-2$ for any sentence.
  This implies that the EOS can be judged easily by whether a particular dimension of 
  $\bm{c}_t$ is approximately $-2$ or not.
  
  During this study, \newcite{Suzgun:19} independently discovered a similar diagram
  as Figure~\subref*{fig:grid} and \subref*{fig:grid-slib}.
  However, their experiments are conducted only on a very simple formal language 
  Dyck-\{1,2\} and the number of dimensions are less than 10, as opposed to
  our experiments in empirical data and high dimensionality of over 100
  on the state vectors.
  
  \subsection{Prediction with a Single Component}
  \label{sec:RepNest}
  
\begin{figure}[!t]
    \centering
    {\tabcolsep=0pt
    \begin{tabular}{llll}
     \raisebox{2.8em}{(a)}~ & 
     \includegraphics[width=0.45\linewidth,trim=0 20 20 10,clip]
       {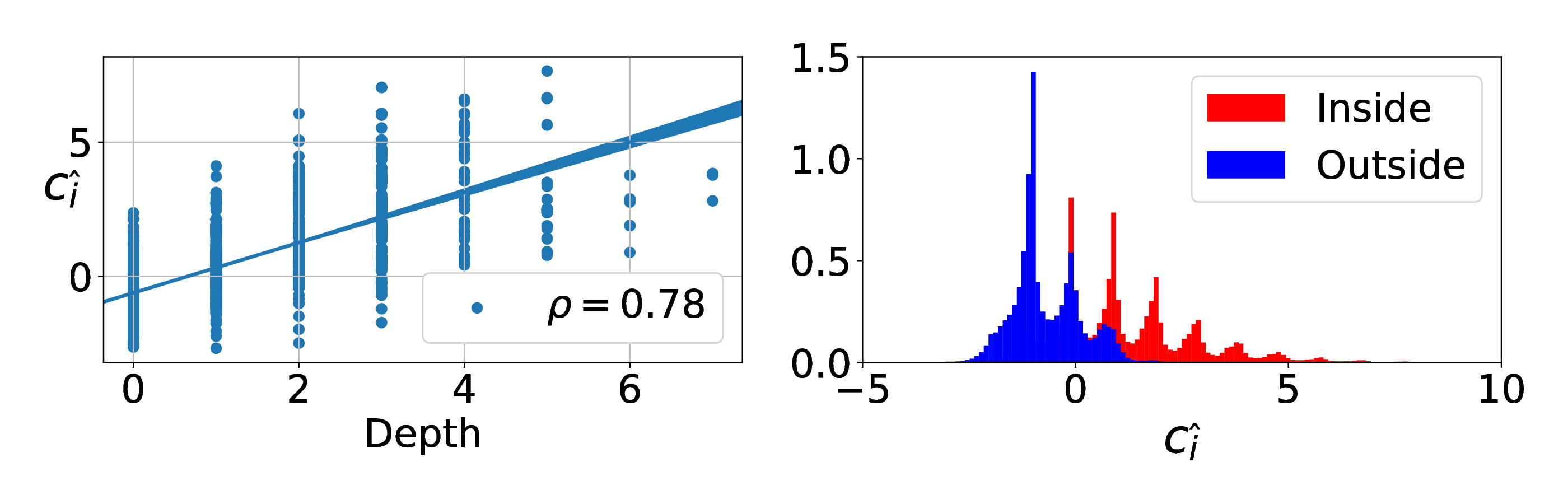} & 
     \raisebox{2.8em}{(b)}~ & 
     \includegraphics[width=0.45\linewidth,trim=20 20 0 25,clip]
       {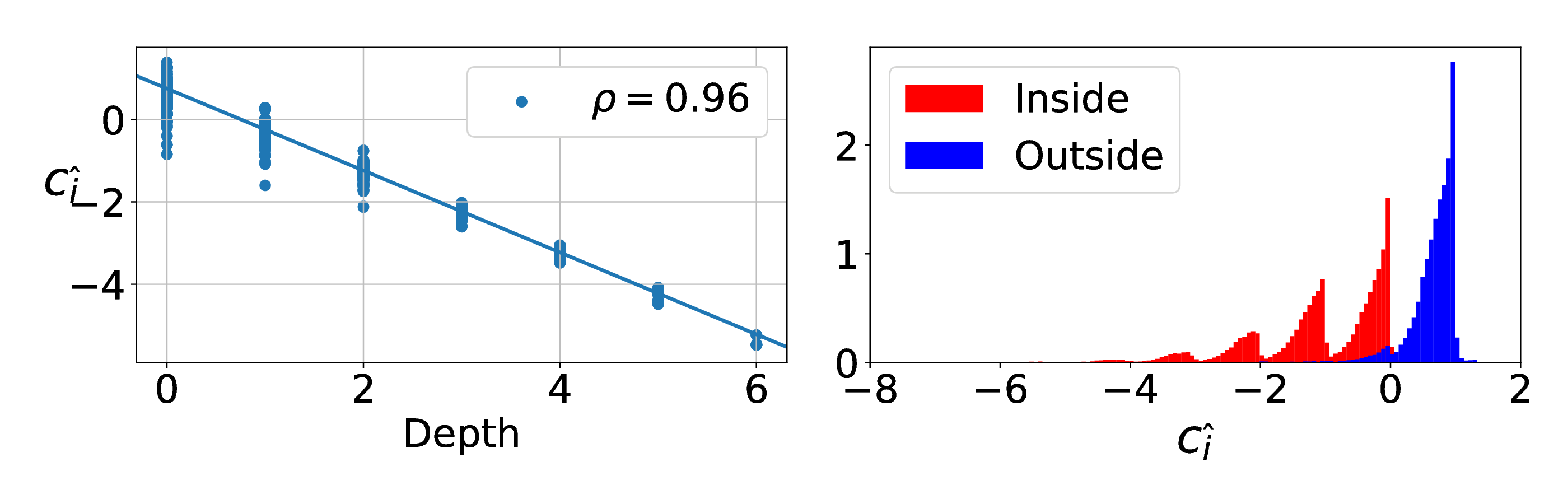} 
     \\
     \raisebox{2.8em}{(c)}~ & 
     \includegraphics[width=0.45\linewidth,trim=10 20 10 25,clip]
       {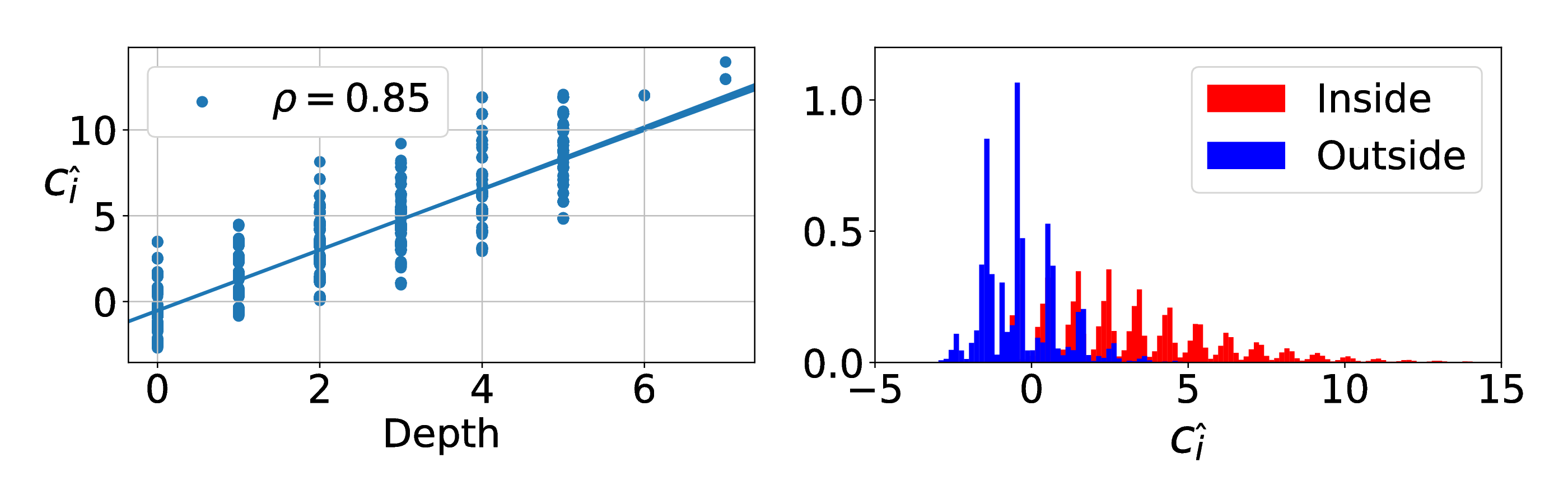} & 
     \raisebox{2.8em}{(d)}~ & 
     \includegraphics[width=0.45\linewidth,trim=20 20 0 25,clip]
       {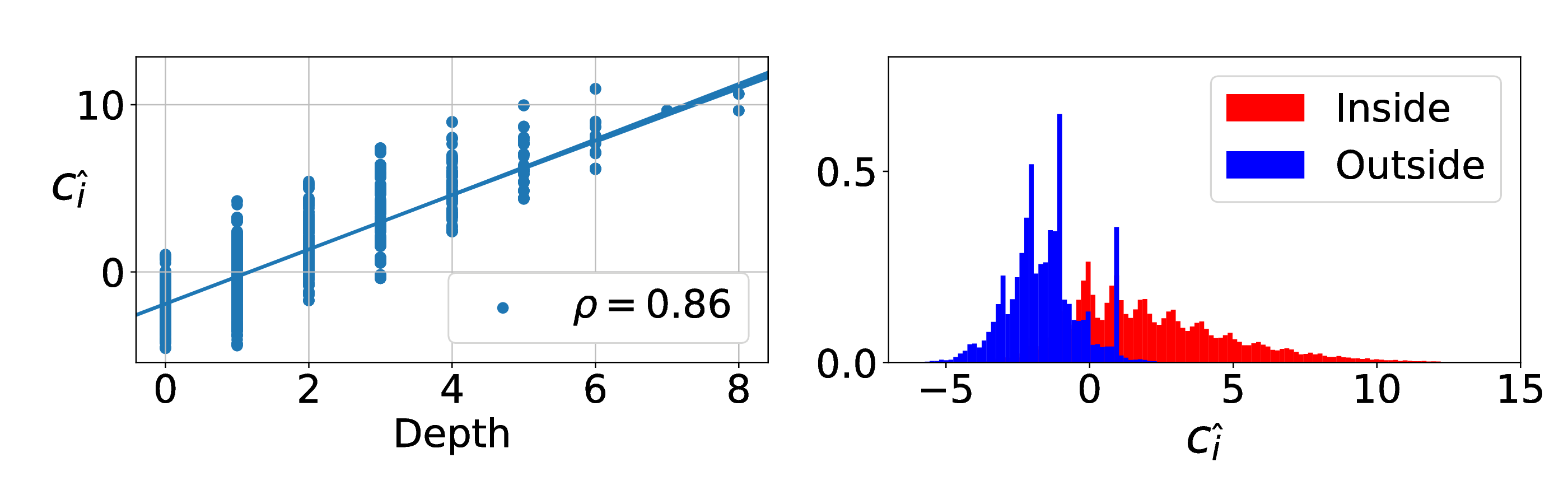} 
     \\
     \multicolumn{4}{c}{ \raisebox{2.8em}{(e)}~
     \includegraphics[width=0.45\linewidth,trim=30 20 0 25,clip]
       {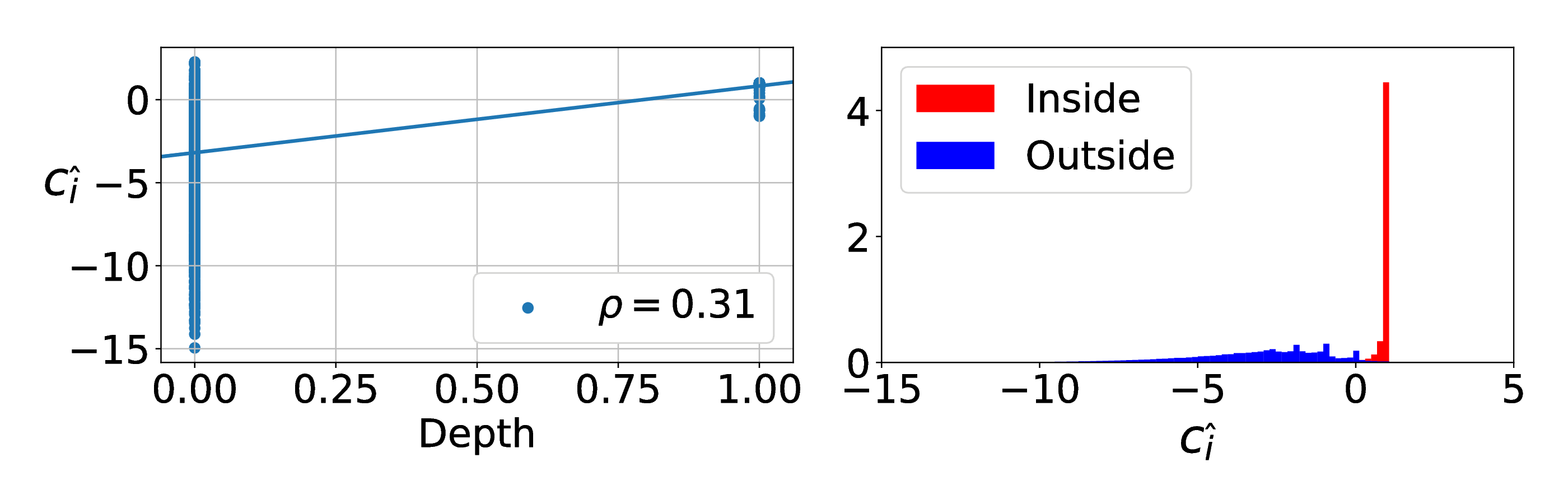}%
     }
    \end{tabular}
    }
    \vspace{-0.5em}
    \caption{Relations between the depth of nesting of phrase structures and characteristic dimensions. 
     $\rho$ denotes the maximum correlation. The target tags are: (a)--(b) NP for {\TG} and {\TGW}, (c)--(d) VP for {\TG} and {\TGW}, (e) NN for {\PNW}.}
     \label{fig:nesting}
  \end{figure}

  \begin{wraptable}[11]{r}[0mm]{70mm}
  \small
      \vspace{-4em}
      \caption{$L_1$ logistic regression from $\bm{c}$ to determine VP for {\TG} and {\TGW}. 
      We show the number of nonzero elements (\#nnz) and its ratio for each regularization.
      The chance level of prediction is around $0.7$.}
      \label{tab:logistic}
      \begin{tabular}{l@{\kern-0.1ex}rr|l@{\kern-0.1ex}rr|c}
      \multicolumn{3}{c}{Dataset {\TG}} \vrule& \multicolumn{3}{c}{Dataset {\TGW}} \vrule& $C$ \\ 
       Acc & \#nnz & ratio & Acc & \#nnz & ratio & {}     \\
      \hline
       {\bf 0.996} & 82 & 41\%  & {\bf 0.9996} & 134 &  13\% & $3\!\times\! 10^{-3}$\rule{0pt}{0.95em} \\
       0.994 & 56 & 28\%  & 0.9992 & 100 &  10\% & $1\!\times\! 10^{-3}$ \\
       0.991 & 34 & 17\%  & 0.998  &  71 &   7\% & $3\!\times\! 10^{-4}$ \\
       0.98  & 21 & 11\%  & 0.991  &  51 &   5\% & $1\!\times\! 10^{-4}$ \\
       0.96  & 8  &  4\%  & 0.97   &  27 & 2.7\% & $3\!\times\! 10^{-5}$ \\
       0.91  & {\bf 5}  &2.5\%  & 0.87   & {\bf 12} & 1.2\% & $1\!\times\! 10^{-5}$ \\
      \hline
      \end{tabular}
  \end{wraptable}
  For {\TG} and {\TGW},
  there are no dimensions that completely correlate with the depth of the nesting 
  unlike {\PN} and {\PNW}.
  We extract a dimension $\hat{\imath}$ that has the largest correlation, and plot the relations between $c_{\hat{\imath}}$ and the depth of the nesting of NP and VP in Figure~\ref{fig:nesting}.
  While the absolute value of $c_{\hat{\imath}}$ increases almost linearly with depth, 
  the variance is not small except for NP on {\TG}.
  Thus, we cannot say that a single element of $\bm{c}$ purely encodes the depth of the nesting
  for a particular tag.
  
  Each of the right half of Figure~\ref{fig:nesting} shows the two histograms that correspond to $c_{\hat{\imath}}$. 
  We can observe that each activation histogram has peaks at integer values. 
  This shows the effect of the natural quantization of $\bm{c}$.
  We call the ratio of the overlap of the normalized histograms as \emph{histogram overlap ratio}.
  The closer the histogram overlap ratio is to $0$, 
  the higher discriminative accuracy of the dimension.
  The minimum histogram overlap ratio of {\TGW} are $0.28$ for VP and $0.06$ for NN.
  From the perspective of histogram overlap ratio, 
  it is easy for NN and slightly difficult for VP to classify whether a word is
  in that phrase by a single dimension.
  
  For NN (common noun) tag,
  from Figure~\subref*{fig:nn}, it can be seen that there are no single dimension in $\bm{c}$ that highly correlate with the depth of the nesting ($\rho\!=\!0.31$).
  On the other hand, the minimum histogram overlap rate is $0.07$, which is sufficiently low. 
  The right histogram of the Figure~\subref*{fig:nn} shows that the occurrence of the token `(NN' has an effect of resetting some dimension of $\bm{c}$.

  \subsection{Representation by a Subspace}
  
\begin{figure}[!t]
   {\small\tabcolsep=0pt
  \centering
   \begin{tabular}{lclc}
    \begin{minipage}[b][30mm][c]{5em} (a) \TG  \end{minipage}
    & \includegraphics[width=0.3\linewidth]{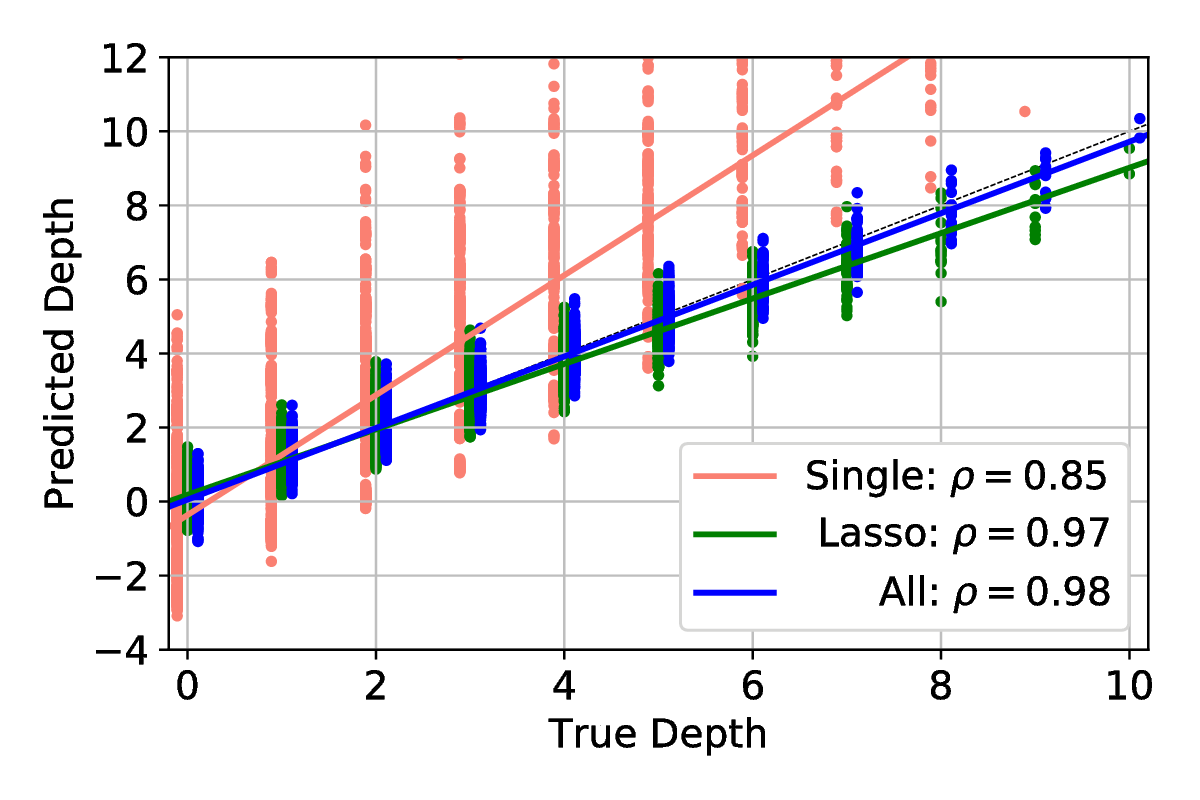}
    & ~~\begin{minipage}[b][30mm][c]{5em} (b) \TGW  \end{minipage}
    & \includegraphics[width=0.3\linewidth]{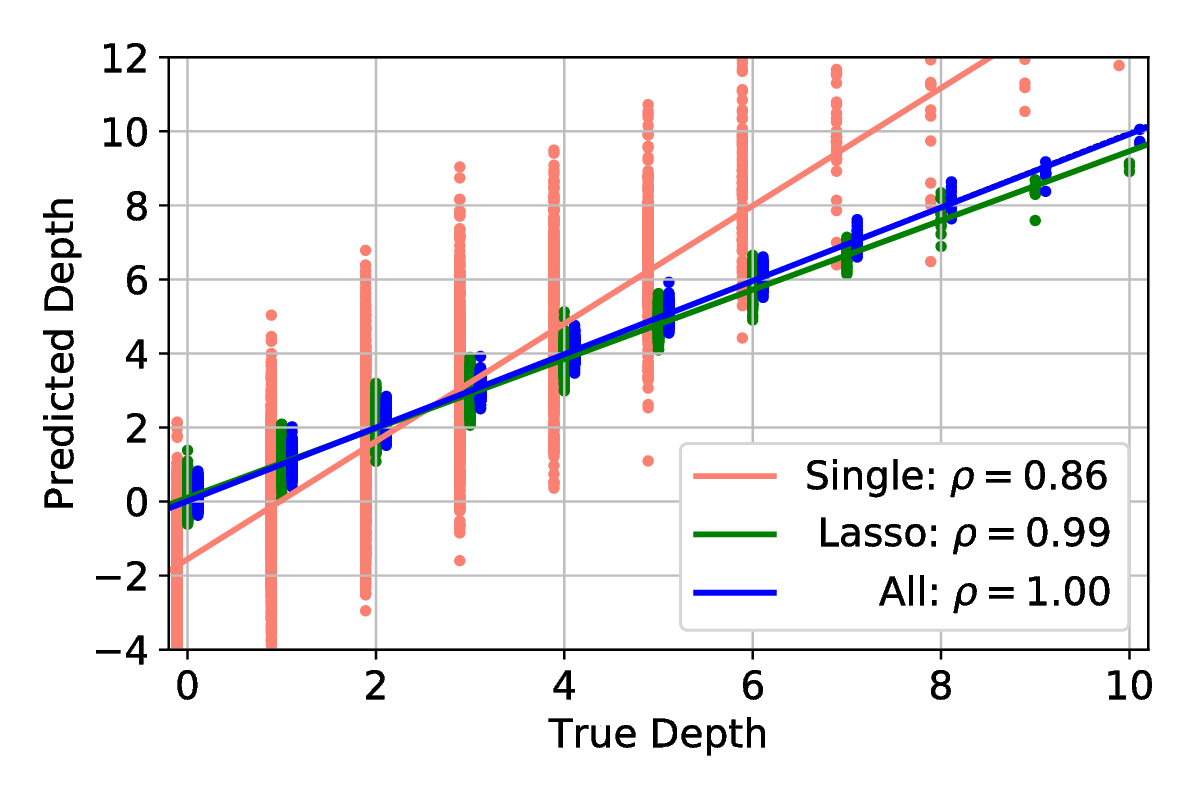}
    \\ \begin{minipage}[b][30mm][c]{5em} (c) \WD \end{minipage}
    & \includegraphics[width=0.3\linewidth]{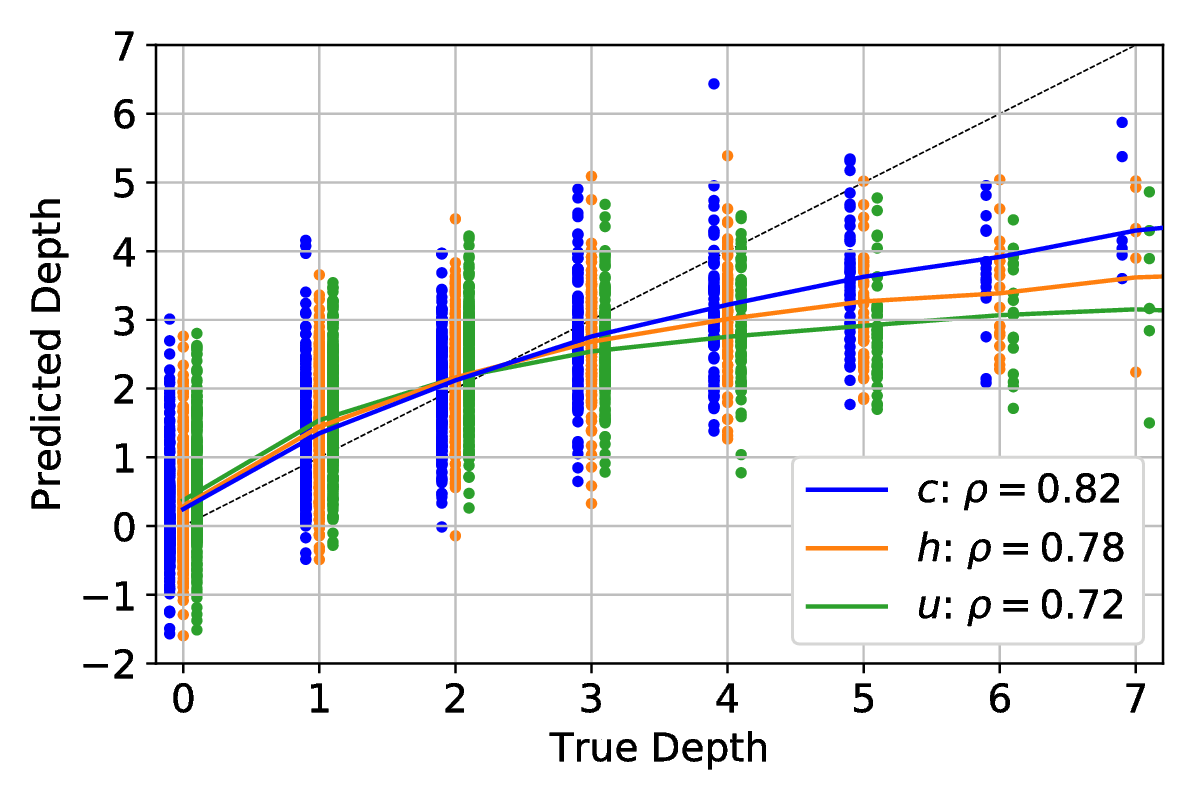} 
    & ~~\begin{minipage}[b][30mm][c]{5em} (d) Lisp \end{minipage}
    & \includegraphics[width=0.3\linewidth, trim=10 10 10 10,clip]{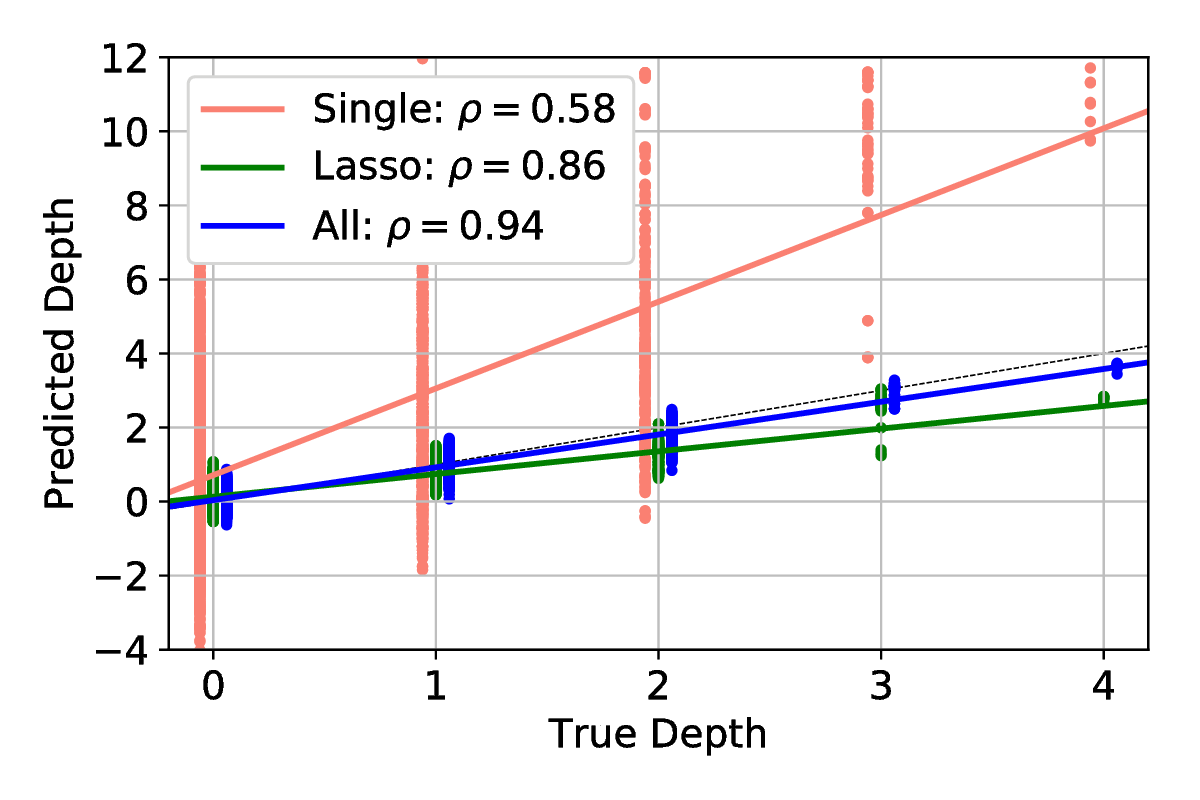} 
   \end{tabular}
   }
   \vspace{-1em}
   \caption{Lasso ($L_1$) regression as a function of the nesting depth of VP for (a)-(c) and {\tt lambda} for (d).
   Linear regression and $c_{\hat{\imath}}$ with the highest correlation coefficient are also shown. 
   The regression results are scaled, and the depth plot is slid slightly shifted to the right for clarity.}
   \label{fig:lasso-VP}
  \end{figure}

  To find a clear representation of the depth of the nesting within $\bm{c}$,
  we try to extract a subspace that have high correlations with it.
  First, we adopt linear regression to predict the depth of nesting from $\bm{c}$. 
  Second, we examine the number of effective dimensions;
  the results of regression for VP are shown in Figure~\ref{fig:lasso-VP}(a)--(c).
  Compared with choosing the best single dimension, the correlation coefficients are clearly improved and almost equals to $1$; $0.983$ for {\TG}, $0.995$ for {\TGW}.
  This also holds for the nesting of {\tt lambda} in Lisp programs where it is $0.940$.
  We also empirically show that a few dimensions are sufficient to classify whether a word is in VP or not.
  Table~\ref{tab:logistic} shows the classification accuracies:
  for {\TGW}, we can keep the accuracy more than $0.99$ while the ratio of non-zero dimensions
  decreases to $5\%$. 
  For the case of {\WD}, {\it i.e.} learning from raw text, the coefficients become smaller but still have
  positively correlate with $\bm{c}$, as shown in Figure~\ref{fig:lasso-VP}(c);
  compared to $\bm{\cupp}$, $\bm{c}$ has the smallest prediction error.
  In summary, the depth of the nesting of phrase structure can be represented by a sum of a relatively small number of elements of the context vector $\bm{c}$, and this relationship is approximately linear.
  The prediction for {\WD} is less accurate than the other datasets with implicitly-given syntax.

  \section{Internal Representation of Syntactic Functions}
  \label{sec:dataset3}
  
  Finally, we investigate how syntactic functions, such as part-of-speech (POS)  and functional words, 
  are represented in internal vectors when LSTM is trained for raw text.
  We also show that their syntactic functions are naturally represented in the context-update vector 
  $\bm{\cupp}$, rather than $\bm{c}$.
  
  \begin{figure}[!b]
  \begin{minipage}[b][65mm][b]{0.25\linewidth}
    \subfloat[][$\bm{c}$]{
     \includegraphics[width=\columnwidth,trim=47 30 25 20,clip]
     {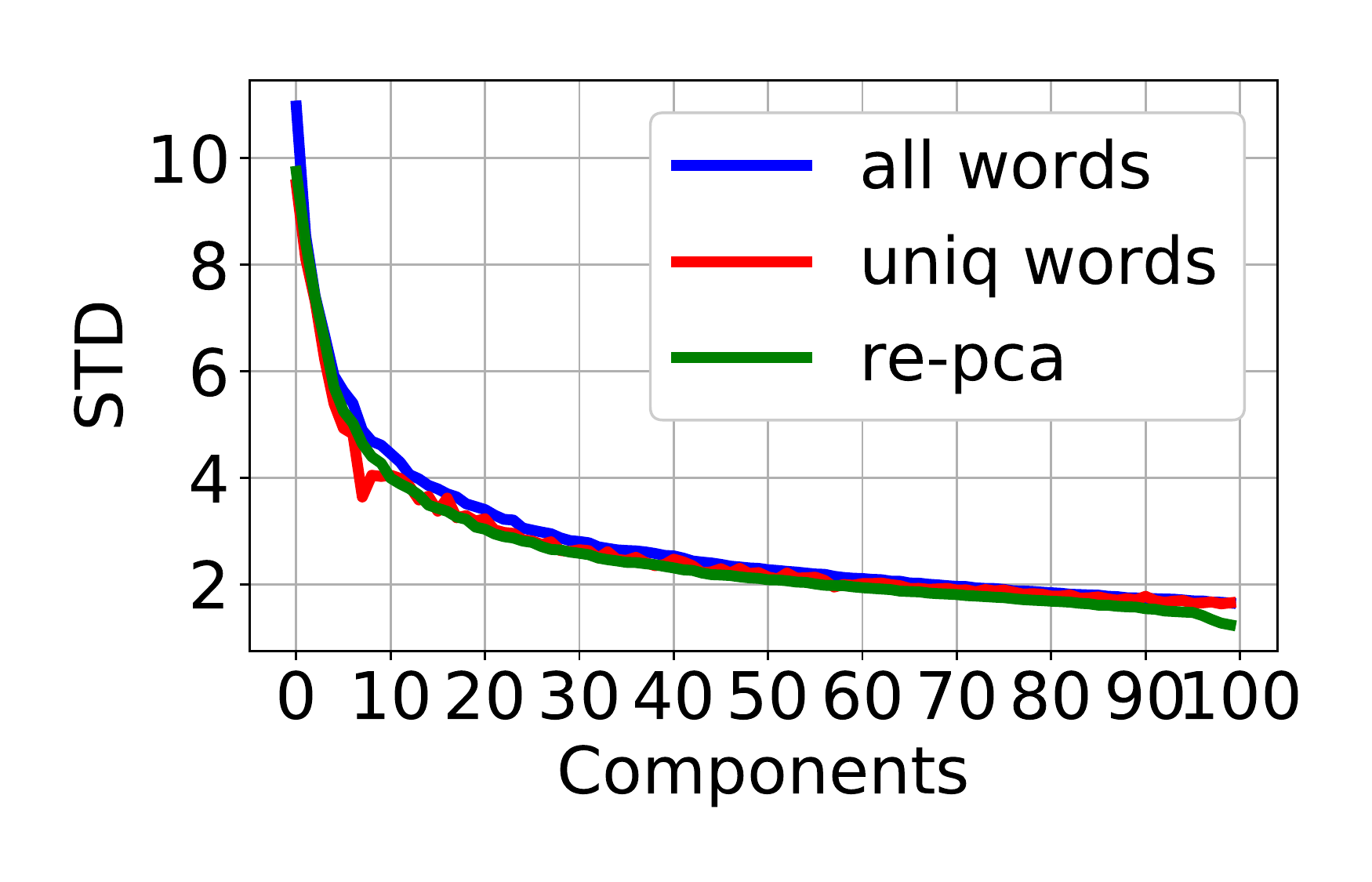}
     \label{fig:cs_pca_plot1}
    }\\
    \subfloat[][$\bm{\cupp}$]{
     \includegraphics[width=\columnwidth,trim=47 30 25 20,clip]
     {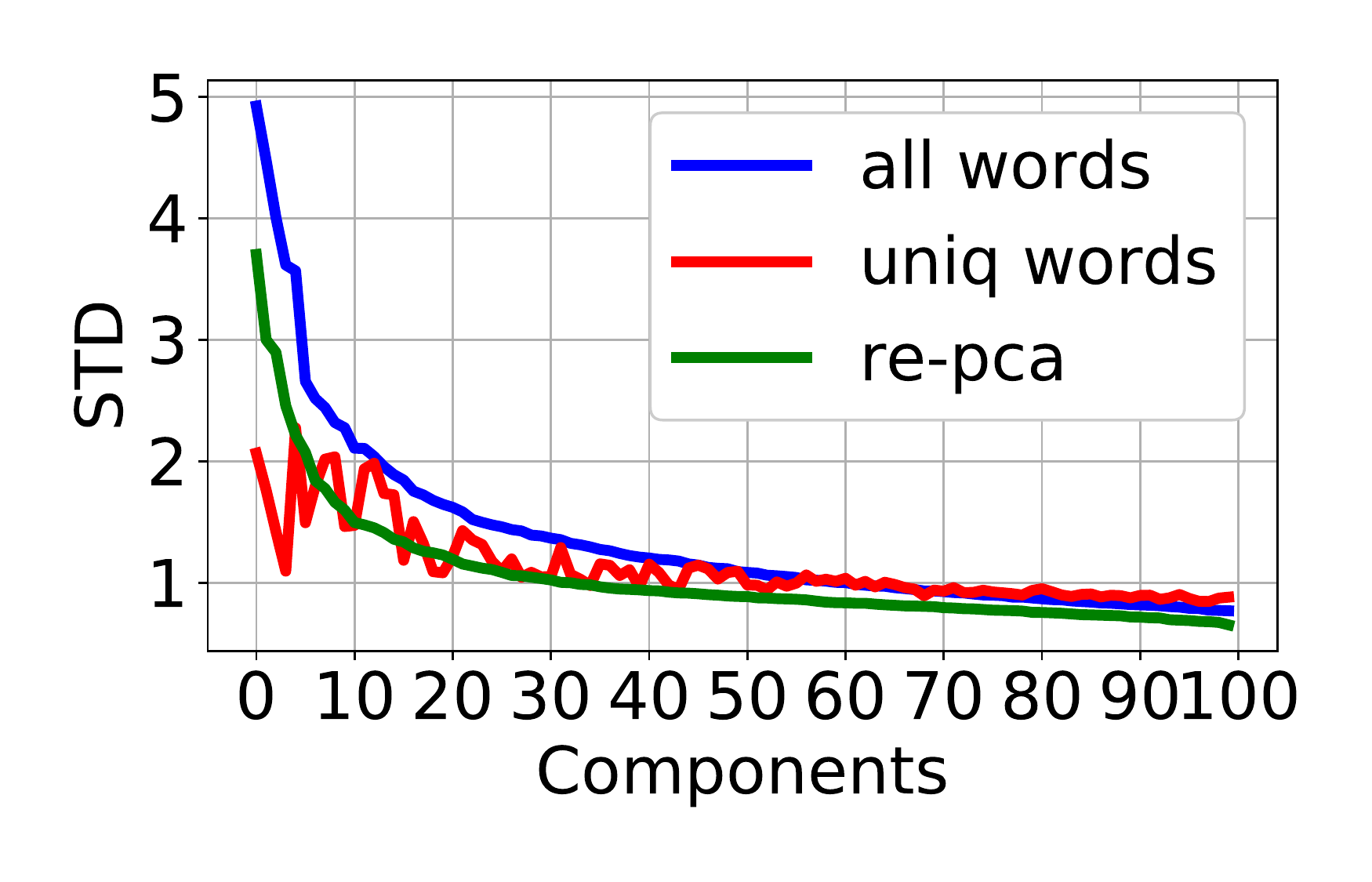}
     \label{fig:cups_pca_plot2}
    }
    \end{minipage}
  \hspace{2mm}
  \begin{minipage}[b][65mm][b]{0.7\linewidth}
   {\tabcolsep=0pt
   \newcolumntype{A}{>{\centering\arraybackslash}p{0.05\textwidth}}
   \newcolumntype{C}{>{\centering\arraybackslash}p{0.19\textwidth}}
   \begin{tabularx}{0.8\textwidth}{ACCCCC}
    & \scalebox{0.95}{VB} & \scalebox{0.95}{VBZ} & \scalebox{0.95}{NN} 
    & \scalebox{0.95}{NNS} & \scalebox{0.95}{CD} \\[-.2em]
    & \scalebox{0.85}{[Verb, base]} & \scalebox{0.85}{[Verb, singular]} 
    & \scalebox{0.85}{[Noun]} 
    & \scalebox{0.85}{[Noun, plural]} & \scalebox{0.85}{[Number]}
   \end{tabularx}
   }
    \\[-0.2em]
    \subfloat[][Primary components for $\bm{\cupp}$ (PCA)]{
       \includegraphics[width=\textwidth,trim=0 30 0 30,clip]{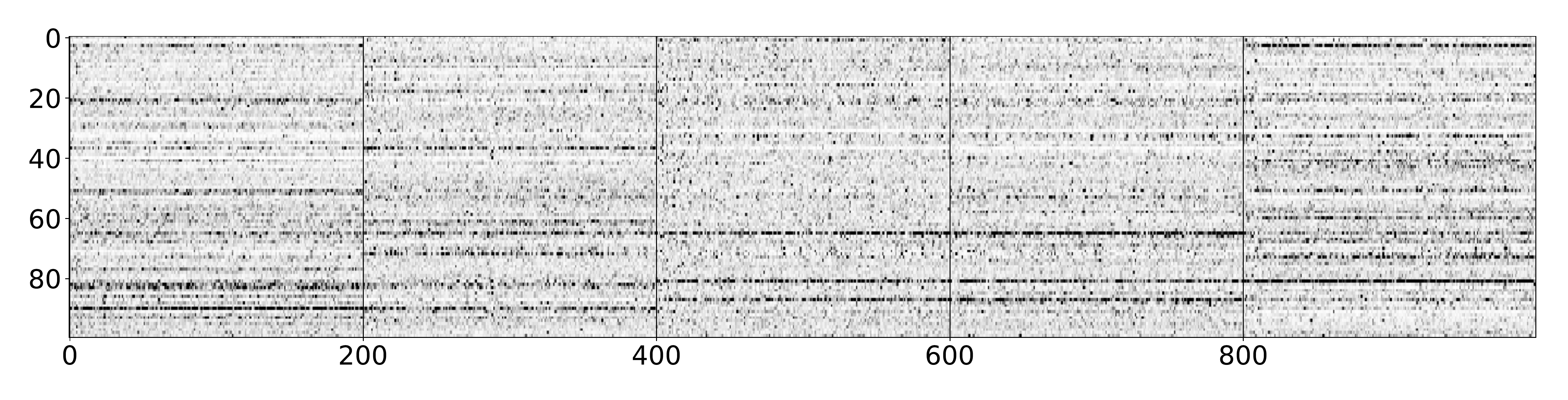}
     } 
     \\[1em]
     \subfloat[][Primary components for $\bm{\cupp}$ (re-PCA)]{
       \includegraphics[width=\textwidth,trim=0 30 0 30,clip]{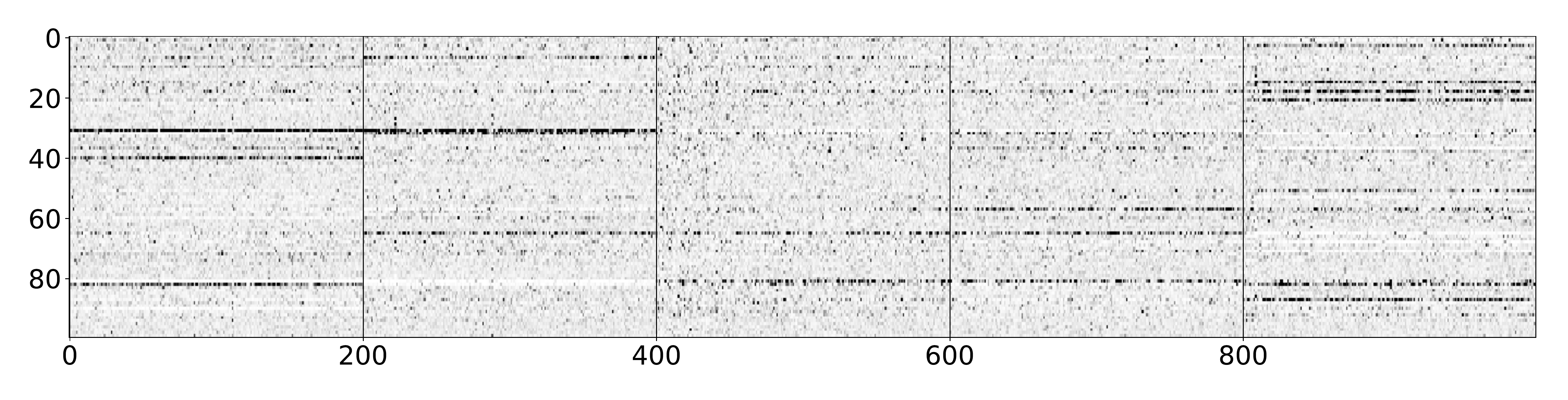}
     } 
   \end{minipage}
   \vspace{-0.5em}
   \caption{
    (a),(b): Distribution of values for each principal component on $\bm{c}$ and $\bm{\cupp}$.
   Vertical axis represents a standard deviation.
   (c),(d): PCA and re-PCA (see text) results of $\bm{\cupp}$ vectors in learned LSTM.
   From left,  VB, VBZ, NN, NNS, and CD, respectively.
   Characteristic dimensions of each POS can be distinguished.
   Note that LSTM is learned from raw text in this scenario.
   \\[-1.5em]
   }
   \label{fig:cs_pca_scatter}
  \end{figure}
  
  \subsection{Representation of a Part-of-Speech}
  
  We investigate whether the LSTM-LM automatically recognizes POS when learning from the raw text,
  because it is difficult to acquire higher phrase structures without ever recognizing POS.
  For this purpose, we employ principal component analysis (PCA) to reduce the dimensionality of
  internal vectors of LSTM to observe unsupervised clusters.
  In Figure~\ref{fig:cs_pca_scatter}-(a)(b),  the vertical axis denotes the standard deviation of
  each principal component over the observed data.
  The statistics over all the occurrences of words represented by the blue line shows
  that the variances are largely influenced by frequent words.
  Therefore,
  next we computed the principal components over unique words,
  as represented by the red lines.
  For $\bm{\cupp}$, the standard deviations for the main components decrease after this processing.
  This implies that the variance within each frequent word significantly affects the result of the PCA.
  
  \subsubsection{Cancelling Frequencies on PCA and Representation in $\bm{\cupp}$}
\begin{figure}[!t]
  \begin{minipage}[b][69mm][t]{0.65\linewidth}
   \tabcolsep=0pt \arrayrulewidth=0.5pt
   \begin{tabular}{c|l}
    \raisebox{3em}{$\bm{c}$}~~ &
    \includegraphics[width=0.96\columnwidth,trim=0 0 20 0,clip]
    {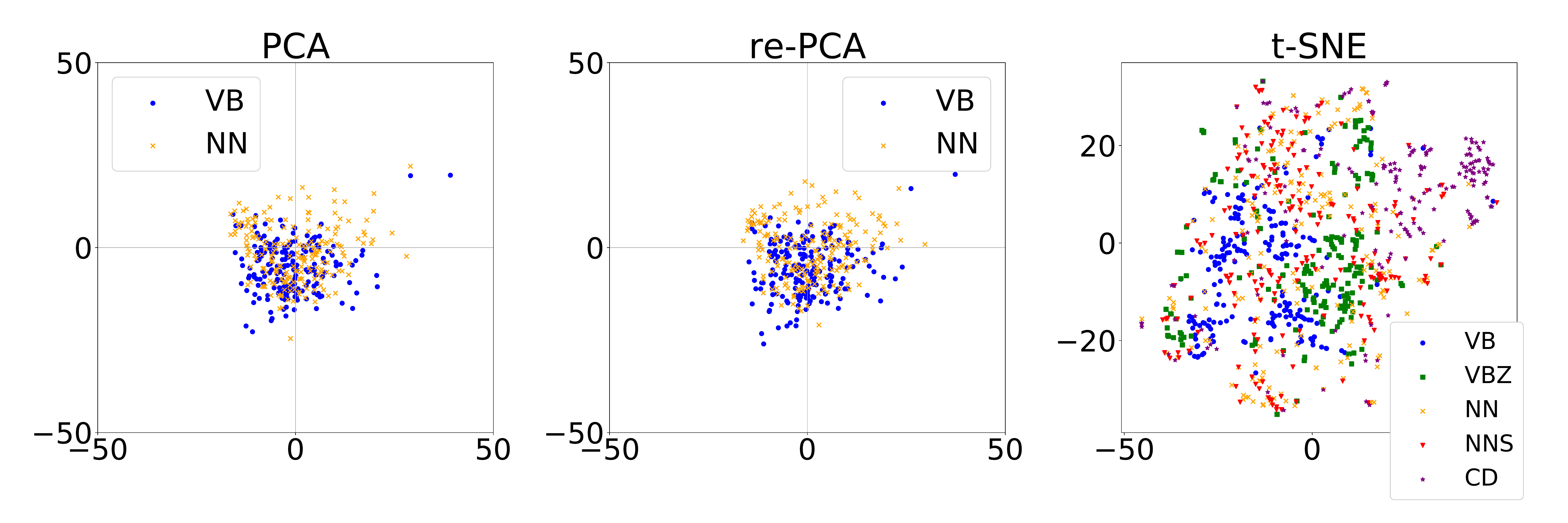}
    \\ \hline
    \raisebox{3em}{$\bm{\cupp}$}~~ &
    \includegraphics[width=0.96\columnwidth,trim=0 0 20 0,clip]
    {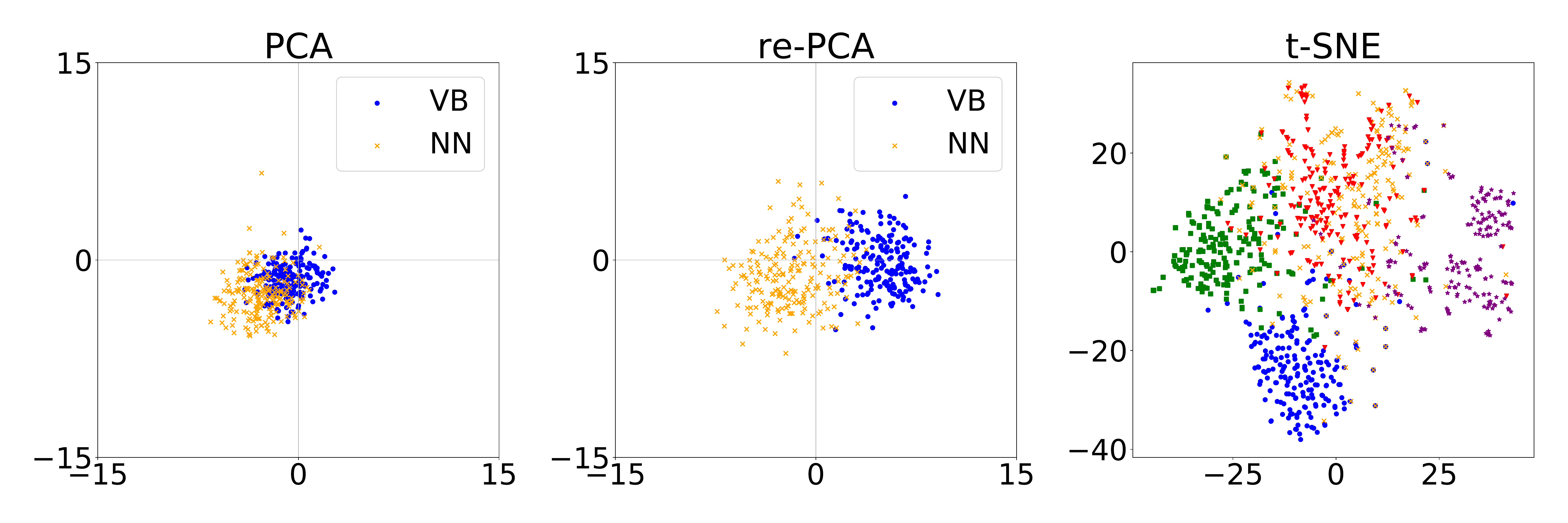}
   \end{tabular}
   \end{minipage}
  \begin{minipage}[b][69mm][c]{0.3\linewidth}
    \includegraphics[width=\linewidth]{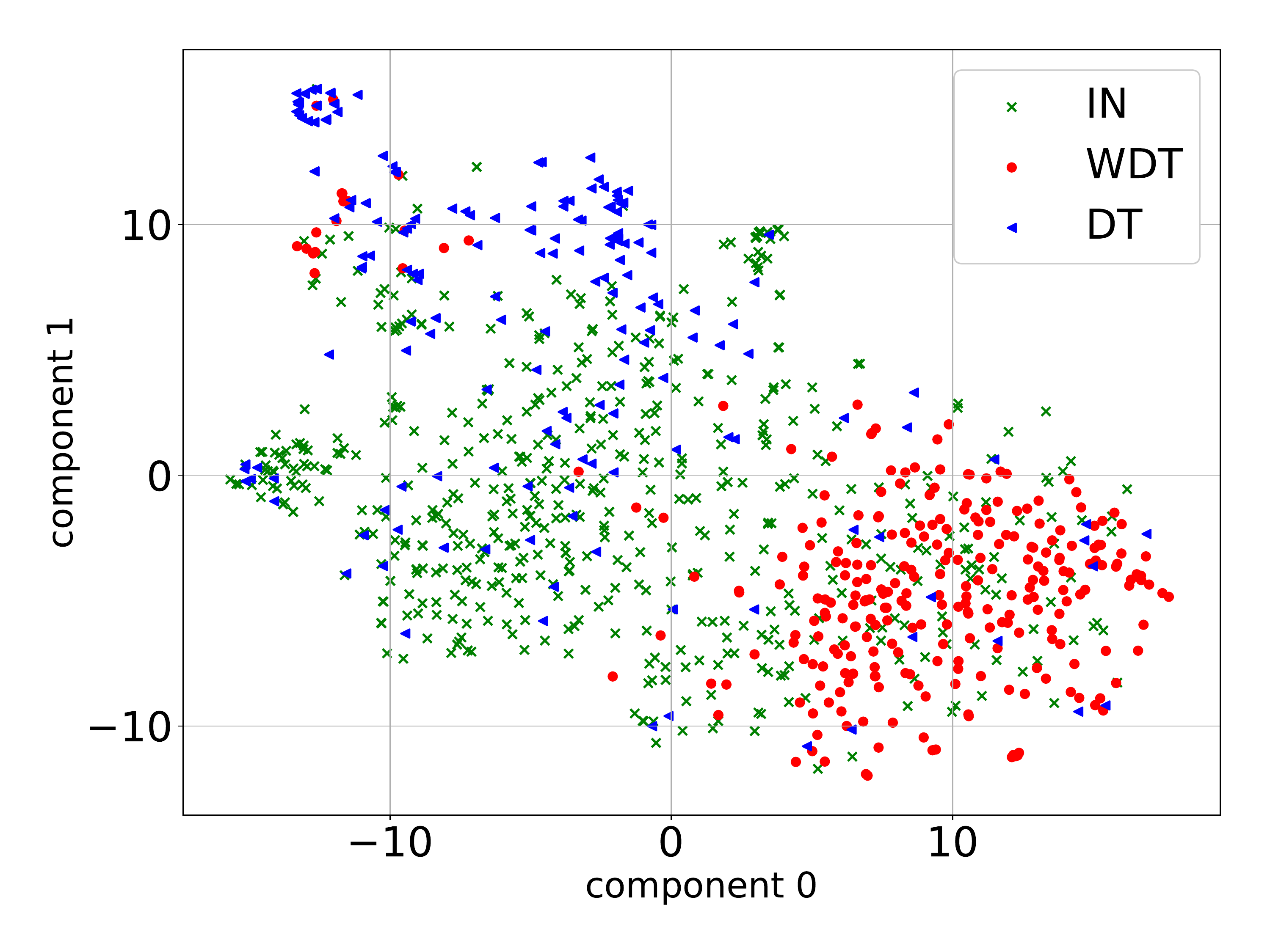}
  \end{minipage}
   \vspace{2em}
    \caption{
    (Left): Result of PCA, re-PCA, and t-SNE applied to the internal vectors: $\bm{c}$ (upper row) and $\bm{\cupp}$ (lower row).
   For PCA and re-PCA, the first and second primary components are shown.
   (Right): Representation of ``that'' in $\bm{\cupp}$ for each usage 
    in the corpus (t-SNE).
    Part of speech (not used in learning) are marked with different colors.
   \\[-1.2em]
   }
   \label{fig:tsne}
   \label{fig:pca2D}
   \vspace{-2em}
\end{figure}

  Figure~\ref{fig:cs_pca_scatter}(a) shows the top 50 components of PCA.
  To enhance readability of both positive and negative values, 
  after the upper-half of the graph ($x$) 
  is negatively copied to ($-x$), each value is filtered by $\exp(\cdot)$
  and shown on the y-axis.
  Figure~\ref{fig:cs_pca_scatter}(b) shows the effect of cancelling the frequencies of the words (\emph{re-PCA}).
  In re-PCA, after PCA is applied to the internal vectors of all the occurrences, it is applied again to
  the averaged vectors, each of which corresponds to each unique word. 
  We can see that POSs are clustered in $\bm{\cupp}$ in an unsupervised fashion. 
  In particular, the result of re-PCA shows there are some dimensions
  that clearly distinguish 
  similar types such as VB and VBZ, NN and NNS, and also between them.
  Furthermore, the distinction between verbs and nouns is evident
  in the first principal component of 
  re-PCA, at the left panel of of Figure~\ref{fig:pca2D}.
  
  \subsection{Representation of Functional Words}
  \label{sec:similarities}
  
  Because functional words play an important role in syntactic parsing, 
  revealing their representation in the internal vectors is important for understanding 
  the mechanism of the syntax acquisition by LSTM.
  To verify if $\bm{\cupp}$ and other internal vectors represent syntactic role of 
  functional words,
  we first take the average of vectors for each word, 
  and compute the cosine similarities between them.
  Table~\ref{tab:similarities} lists words that have the highest similarities to some instances of words.
  From the tables, it can be seen that the context-update vector $\bm{\cupp}$ 
  captures their syntactic role more appropriately than the context vector
  $\bm{c}$ itself.
  Since $\bm{c}$ possesses contextual information in a sentence, 
  the co-occurrence of words will affect the similarity through $\bm{c}$.
  We also examined $\bm{h}$ and confirmed that its clustering ability is basically similar to $\bm{c}$. 
  
  \begin{table}[t]
  {\footnotesize \tabcolsep=0pt
  \caption{Similarities among internal vectors for several functional words. $\bm{c}$ and $\bm{\cupp}$ are averaged over occurrences of each word. }
  \label{tab:similarities}
  \begin{tabular*}{0.24\columnwidth}{@{\extracolsep{\fill}}lclc}
  \multicolumn{4}{l}{``her''} \\
          $\bm{c}$ &  sim. &    $\bm{\cupp}$ &  sim. \\
  \hline
        his &  0.70 &    his &  0.39 \\
     mother &  0.68 &     my &  0.33 \\
    playing &  0.67 &    the &  0.28 \\
       mind &  0.66 &    its &  0.26 \\
    husband &  0.65 &    our &  0.26 \\
    matters &  0.65 &   your &  0.26 \\
      party &  0.65 &  their &  0.25 \\
  \hline
  \end{tabular*}
  ~
  \begin{tabular*}{0.24\columnwidth}{@{\extracolsep{\fill}}lclc}
  \multicolumn{4}{l}{``his''} \\
        $\bm{c}$ &  sim. &    $\bm{\cupp}$ &  sim. \\
  \hline
        the &  0.74 &    the &  0.43 \\
         's &  0.73 &  their &  0.39 \\
          a &  0.72 &    her &  0.39 \\
      their &  0.71 &   your &  0.37 \\
          ' &  0.71 &    its &  0.37 \\
        her &  0.70 &      a &  0.36 \\
        its &  0.70 &     's &  0.36 \\
  \hline
  \end{tabular*}~
  \begin{tabular*}{0.24\columnwidth}{@{\extracolsep{\fill}}lclc}
  \multicolumn{4}{l}{``an''} \\
               $\bm{c}$ &  sim. &         $\bm{\cupp}$ &  sim. \\
  \hline
               a &  0.71 &           a &  0.31 \\
             the &  0.68 &         the &  0.27 \\
         initial &  0.68 &         its &  0.26 \\
        enormous &  0.67 &     another &  0.25 \\
   \scalebox{0.9}{opportunity} &  0.67 &  her &  0.25 \\
         planned &  0.67 &         any &  0.25 \\
        military &  0.66 &         his &  0.22 \\
  \hline
  \end{tabular*}
  ~
  \begin{tabular*}{0.24\columnwidth}{@{\extracolsep{\fill}}lclc}
  \multicolumn{4}{l}{``a''} \\
                   $\bm{c}$ &  sim. &      $\bm{\cupp}$ &  sim. \\
  \hline
              the &  0.76 &      the &  0.43  \\
           modest &  0.76 &  another &  0.36  \\
               's &  0.75 &      his &  0.36  \\
               to &  0.74 &     your &  0.34  \\
              its &  0.73 &       's &  0.33  \\
          similar &  0.73 &    every &  0.33  \\
              and &  0.73 &      its &  0.33  \\
  \hline
  \end{tabular*}
  }
  \end{table}
  
  \subsection{Representation of Ambiguity with Functional Words}
  
  A word ``that'' is a representative ambiguous functional word that has multiple grammatical meanings:
  it has three main meanings that are syntactically similar to the words 
  ``{\it if}'', ``{\it this}'', and ``{\it which}'', respectively. 
  Figure~\ref{fig:tsne} shows how these meanings are encoded in $\bm{\cupp}$, by mapping 
  to two dimensions using t-SNE.
  Although they are not completely separated, we can see that
  they are clustered according to their syntactic behaviors in context.
  
  \section{Related Work}
  \label{sec:related-work}
  As research on how LSTM tracks long-term dependence,
  behaviors of LSTM with several dimensions 
  have been studied using artificial languages \cite{Tomita:82,Prez-Ortiz:03,Schmidhuber:15}.
  With recent applications of LSTM to various tasks,
  studies are being conducted on how LSTM recognizes syntax and long-term dependencies 
  \cite{Adi:17,Li:16}.
  For instance, Linzen et al.~\shortcite{Linzen:16} uses number agreement 
  to determine whether a language model using LSTM truly captures it.
  Khandelwal et al.~\shortcite{Khandelwal:18} evaluates how the distance between words affects the prediction in LSTM-LM.
  Weiss et al.~\shortcite{Weiss:18b} utilize the learned LSTM to 
  construct deterministic automata.
  Furthermore, Avcu et al.~\shortcite{Avcu:17} control the complexity of long-range dependency using SP-$k$ languages, and verify if LSTM can track them.
  Several studies have attempted to theoretically understand the learning ability of language models
  using RNNs, including LSTM and GRU~\cite{Cho:14,Chen:18,Weiss:18a}.

  
  \section{Conclusion}
  
  In this paper, we empirically investigated various behaviors of LSTM on natural text
  by looking into its hidden state vectors.
  Contrary to previous work that deal with only artificial data,
  we clarified that updates $\bm{u}$ of the context vectors $\bm{c}$ are approximately
  discretized and accumulated in a low-dimensional subspace,
  leading to an approximate counter machines discussed in Section~4 and
  a clear representation of syntactic functions as shown in 
  Section~5, in spite of the high dimensionality of state vectors
  explored in this study.
  Especially, we show that the representations of POS are acquired in the space of $\bm{u}$ rather than $\bm{c}$ and $\bm{h}$ in an unsupervised manner. 
  The fact that the first principal component of re-PCA for $\bm{u}$ encodes the difference between NP and VP
  is not only significant for understanding how LSTM-LMs acquire syntactics, but also regarded as a result of extracting the most important syntactical factor 
using LSTM with respect to the target language.




\addcontentsline{toc}{section}{References}
\bibliographystyle{coling}
\bibliography{coling2020}

\end{document}